# MAGIC-SSCIL: Manifold Anchoring and Geometric Incremental Calibration for Semi-Supervised Class-Incremental Learning

Yousef Abdi [a], Mohammad Asadpour [a*], Yousef Seyfari [b]

## Abstract

Exemplar-free semi-supervised class-incremental learning (SSCIL) asks a model to learn new classes from a few labeled and many unlabeled images without storing past data. A common recipe replays old classes from stored Gaussian statistics and pseudo-labels unlabeled data with a confidence threshold. With one or two labels per class, however, the model's confidence is a poor guide to which pseudo-labels are right, so the statistics that replay depends on rest on few or noisy samples. We propose MAGIC (Manifold Anchoring and Geometric Incremental Calibration), which propagates labels over a k-nearest-neighbor graph of frozen features and uses the propagated confidences as weights in the unlabeled loss and, through Soft-Weighted Geometry Calibration (SWGC), in the class statistics from which phantom features replay old classes. With a frozen ResNet-18 on CIFAR-100, CUB-200, and ImageNet-R at 1%, 5%, and 10% labels, we compare MAGIC with nine supervised class-incremental methods equipped with FixMatch and two native SSCIL methods. MAGIC gains from unlabeled data in all nine settings, most at 1%, and leads on fine-grained CUB-200 at every ratio; at 1% it is level with the best on ImageNet-R and 2.2 points behind the native SSCIL method TACLE on CIFAR-100. Experiments inside MAGIC show that most of this gain flows through the weighted statistics, not the loss, that at 1% labels graph pseudo-labels are more accurate than the model's own at equal coverage, and that a fixed threshold of 0.95 removes most of the gain. At higher ratios, baselines that adapt the backbone or model full covariances overtake MAGIC on CIFAR-100 and ImageNet-R.

**Keywords:** Continual learning, Class-incremental learning, Catastrophic forgetting, Semi-supervised learning, Label propagation

## 1. Introduction

Building AI systems that learn continuously, absorbing new experiences without discarding previously acquired knowledge, remains a long-standing open problem in machine learning. This capability, known as continual or lifelong learning, is critical for real-world applications with shifting data distributions over time, such as personalized assistants, autonomous navigation, and clinical decision support. Deep neural networks struggle with these shifts. When training moves on to new tasks, performance on earlier ones tends to collapse abruptly, a phenomenon known as catastrophic forgetting [15, 37]. At its core the problem is a tension between two competing demands: plasticity, the capacity to acquire new information, and stability, the capacity to retain what has already been learned.

[*] Corresponding author: Mohammad Asadpour (m_asadpour@tabrizu.ac.ir)
Yousef Abdi (y.abdi@tabrizu.ac.ir)
Yousef Seyfari (seyfari@maragheh.ac.ir)

[a] Faculty of Electrical and Computer Engineering, University of Tabriz, Tabriz, Iran
[b] Faculty of Engineering, University of Maragheh, Maragheh, Iran

Continual learning is commonly organized into three scenarios [47]. In Task-Incremental Learning (Task-IL), a task identifier is available at inference, which allows task-specific components. Domain-Incremental Learning (Domain-IL) deals with shifts in the input distribution while the label set stays fixed. The most demanding and realistic setting is Class-Incremental Learning (CIL), in which a model learns new, non-overlapping classes sequentially and must maintain a single unified classifier over all classes seen so far, without task identifiers at inference. This matches most real deployments, where new categories keep appearing, whether as user-defined labels, new product types, or previously unseen species in ecological monitoring. It is also where forgetting is most severe, because the model must distinguish old and new classes simultaneously without task identifiers.

Three principal families of methods mitigate forgetting in CIL: 1) Regularization-based approaches constrain updates to parameters identified as important for earlier tasks [29, 32, 57], but their effectiveness degrades over long task sequences as constraints accumulate. 2) Architecture-based methods isolate parameters or dynamically expand capacity for new tasks [43, 56], which protects old knowledge but allows the model to expand without bound and complicates the architecture. 3) Replay-based methods store a small set of past exemplars and revisit them during later training, directly addressing the distribution shift underlying forgetting [24, 42]. This strategy remains a dominant approach, keeping a balanced representation of old and new classes.

Despite their strong empirical performance, replay-based approaches raise serious practical concerns: storing raw exemplars often conflicts with privacy regulations such as GDPR and HIPAA, and proves infeasible on memory-constrained edge devices. To sidestep these issues, exemplar-free continual learning has gained traction. Its goal is to protect previously acquired knowledge without saving any actual training examples, often through latent replay [64], compact prototypes [65], or generative alternatives [44]. Streaming scenarios introduce an additional hurdle: labeled data are often extremely scarce. When these two conditions coincide, we arrive at the setting of Exemplar-Free Semi-Supervised CIL (SSCIL). In this setting, the learner has access only to a sparse stream of labeled instances alongside plentiful unlabeled ones, and must not retain any storage of earlier inputs.

Research on SSCIL is still in its infancy. Building on the pre-trained backbones common in exemplar-free CIL, methods such as TACLE [27] combine two parts: confidence-thresholded pseudo-labeling, which exploits the abundant unlabeled stream, and replay of compact per-class statistics for the old classes. Under sparse supervision each part has a weakness, and the two weaknesses compound each other.

The first weakness concerns sample selection. A confidence threshold decides, sample by sample, whether a pseudo-label is accepted or discarded. When each class has only one to five labeled images, the model is uncertain about most unlabeled samples, including many that it would label correctly. A high threshold then accepts only a small fraction of the stream, while a lower one admits more samples but also more wrong labels. Worse, at the smallest budgets the model's confidence is a poor guide to correctness: on CUB-200 with one label per class, about 70% of the pseudo-labels that a self-trained model accepts at a threshold of 0.95 are wrong (Section 5.4). In practice the threshold is therefore set per dataset and even per task: TACLE, for example, lowers its threshold from 0.90 toward 0.65 over the task sequence [27]. More fundamentally, a threshold on the model's own confidence judges each sample in isolation. A sample with low confidence may sit inside a dense cluster of correctly labeled neighbors, yet it is discarded, and with it the structural information that the unlabeled stream was meant to provide.

The second weakness concerns the class statistics on which replay depends. Methods based on latent or prototype replay estimate each class's mean and covariance during the task in which the class appears and reuse them for the rest of the stream. With one or two labeled samples per class, the mean is unreliable and the covariance is ill-conditioned or singular. Pseudo-labeled samples could fill this gap, but at these budgets a threshold admits either few of them or many wrong ones. The two weaknesses therefore compound: however the threshold is set, the statistics that every later task replays rest on few or noisy samples. If the backbone is also adapted during training, the stored statistics drift out of alignment with the current feature space, and the phantom features drawn from them no longer represent their classes.

To address both weaknesses, we propose MAGIC (Manifold Anchoring and Geometric Incremental Calibration), an exemplar-free framework that replaces the accept-or-reject decision with continuous, neighborhood-based reliability weights. MAGIC assigns unlabeled samples by graph-based label propagation: the labels of the few labeled samples, together with the stored means of the old classes, diffuse over a k-nearest-neighbor graph, so a sample's assignment reflects its neighbors and not only its own confidence. No sample is discarded; each one is weighted by its squared propagated confidence. Before the training of each task, labels are propagated over the frozen-backbone features to produce soft targets for the unlabeled data. At the end of each task, a second propagation over the adapted features, which separate classes better, produces the weights used for estimating class statistics. Soft-Weighted Geometry Calibration (SWGC) then computes the mean and variance of each class from all samples of the task, weighted by these reliabilities, on the frozen-backbone features, so the stored statistics do not drift while the adapter keeps learning.

The rest of MAGIC is deliberately simple. Phantom features sampled from the stored Gaussians replay the old classes through a residual adapter and a cosine classifier on top of the frozen backbone.

Our principal contributions are summarized as follows.

- **How unlabeled data helps statistics-based SSCIL.** Using MAGIC as a controlled testbed, we show that at low label budgets most of the benefit of the unlabeled stream reaches the model through the class statistics used for replay, not through the unlabeled loss. Used only in the statistics, the pseudo-labels keep 63–85% of the gain; used only in the loss, they keep at most 39%, and on CIFAR-100 at 1% labels they even raise forgetting from 23 to 40 points. We also show why fixed thresholds fail at these budgets: the model's confidence is a poor guide to which pseudo-labels are right, whereas the graph's confidence ranks them well, and a threshold of 0.95 keeps only 0.2 of the 14.8 accuracy points that unlabeled data add on CUB-200 at 1% labels.

- **Soft-Weighted Geometry Calibration (SWGC).** We replace confidence thresholding with graph-based label propagation [26, 59] and use its soft assignments to weight every unlabeled sample, both in the training loss and in the estimation of class means and variances. For the statistics, the weights come from a second propagation over the adapted features, which separate classes better, while the statistics themselves are computed on the frozen features, which do not drift.
- **A matched-backbone benchmark with a gain analysis.** We compare MAGIC with nine supervised CIL methods equipped with FixMatch [45] and two native SSCIL methods [11, 27] on CIFAR-100, CUB-200, and ImageNet-R at 1%, 5%, and 10% labels, and we measure how much each method gains from unlabeled data. Confidence-thresholded pseudo-labeling helps, hurts, or does nothing depending on the host method, whereas MAGIC gains in all nine settings, most at 1%. At this budget MAGIC is the best method on CUB-200, level with the best on ImageNet-R and second to TACLE on CIFAR-100, and it has the highest mean accuracy on CUB-200 at every label ratio.

## 2. Related work

Continual learning has been studied extensively [9, 51], but most prior work targets task-aware setups such as Task-Incremental Learning rather than the more demanding class-incremental setting. This section surveys the landscape relevant to Class-Incremental Learning (CIL), beginning with the general problem and narrowing to the specific setting of Semi-Supervised CIL (SSCIL). We discuss strategies for mitigating catastrophic forgetting, the incorporation of unlabeled data, and the emerging role of geometric and structural reasoning.

### 2.1. Class-incremental learning: from replay to exemplar-free strategies

Replay is effective at reducing forgetting in CIL, but storing raw exemplars comes with ongoing storage costs and privacy risks. This makes it unsuitable for settings with strict data governance or limited on-device memory [36, 65]. As a result, a growing body of work now focuses on exemplar-free methods, which preserve past knowledge without keeping any raw data.

Exemplar-free strategies can be categorized by the mechanism they use to approximate lost data. Regularization-based methods such as EWC [29] constrain weight updates to protect parameters that were important for previous tasks. Knowledge-distillation techniques [23, 32] preserve the functional mapping of the old model by distilling its outputs or intermediate representations. A more recent and potent trend exploits the transferability of large-scale pre-trained models. Building on a robust, frozen feature extractor reframes CIL from a problem of joint feature-classifier plasticity into one of incremental classifier alignment within a stable representational space [38, 58, 62].

This paradigm spans a spectrum of backbone plasticity. At the training-free end of the spectrum, a classifier is built directly on frozen features. Several methods follow this approach: SimpleCIL [62] forms nearest-class-mean prototypes, APER [62] adds lightweight first-task adaptation before merging prototypes, FeTrIL [41] synthesizes pseudo-features for old classes by geometrically translating new-class features, FeCAM [19] models each class as a shrinkage-regularized Gaussian under a Mahalanobis metric, and RanPAC [38] applies a frozen random projection to decorrelate class prototypes. At the other end, a small amount of plasticity is retained. For instance, SLCA [58] slowly fine-tunes the backbone while aligning classifiers, as detailed below.

Among these, SLCA (Slow Learner with Classifier Alignment) [58] establishes a strong benchmark. The method proceeds in two stages: it first adapts the pre-trained backbone to the new task, then aligns all classifiers, both old and new, using class-wise mean and variance statistics computed in the feature space. This statistical alignment serves as a form of generative rehearsal, simulating the distribution of past classes without storing exemplars. It exemplifies a broader line of exemplar-free methods that summarize each class through low-order statistics or prototypes [19, 64, 65]. EASE [60] trains a lightweight adapter for each task, so that every task has its own feature subspace, and rebuilds the prototypes of old classes in the new subspaces from their semantic similarity to new classes, without storing exemplars. PROOF [63] carries this idea to vision-language models: it learns task-specific projections on frozen image and text encoders and fuses visual and textual features. MAGIC shares their reuse of a frozen pre-trained representation and their reliance on class prototypes or statistics, but targets the semi-supervised setting, in which those statistics must themselves be estimated from a few labels and many unlabeled images.

SLCA's success underscores the promise of pre-trained models for exemplar-free CIL, but its performance rests on a critical assumption: sufficient labeled data per new class to estimate first- and second-order statistics reliably. In practice, this assumption is often violated because labeling is costly and new tasks frequently arrive with only sparse supervision. Moreover, even slight fine-tuning of the feature extractor on limited labeled data introduces representation drift that misaligns the feature space on which old classifiers depend. The challenge therefore shifts from purely exemplar-free to exemplar-free and label-efficient—a substantially harder setting, and one that demands a semi-supervised approach.

### 2.2. Semi-supervised class-incremental learning (SSCIL)

SSCIL addresses the twin challenges of sequential learning and label scarcity, where the model must learn from a stream of sparse labeled data and abundant unlabeled data. Existing work falls into four broad families.

1. **Generative replay.** Early methods such as ORDisCo [50] and adversarial autoencoder-based approaches [31] attempt to learn the joint distribution of labeled and unlabeled data. However, training generative models incrementally is notoriously unstable and computationally expensive, and often suffers from mode collapse in complex visual domains.

2. **Prototype and distribution modeling.** To avoid the cost of generative modeling, Semi-IPC [34] relies on contrastive pre-training to stabilize the feature space and then builds an incremental prototype classifier with unsupervised regularization. Under label scarcity this yields strong accuracy. In the few-shot setting (SS-FSCIL), parametric approaches offer an alternative: rather than model the raw data, they represent each class by a distribution. k-PPCAs [20], for example, use a mean vector and a shared or learned covariance. Adding a new class then reduces to estimating these statistics, without retraining the backbone. A related idea tackles the drift that a moving backbone causes in stored prototypes. Learnable Drift Compensation (LDC) [18] learns a linear projector that maps old-task features into the current feature space, thereby correcting prototype positions without needing exemplars. All these approaches, like SLCA, depend on the fidelity of the estimated statistics. Under sparse supervision that dependence makes them fragile. LDC partly addresses this issue by correcting the prototype positions instead of re-estimating them.

3. **Pseudo-labeling and consistency.** The most common approach adapts semi-supervised techniques to the incremental setting, in particular FixMatch-style pseudo-labeling and consistency regularization [45]. Here an important distinction arises. Methods that directly integrate FixMatch inherit its fixed confidence threshold. In contrast, dedicated SSCIL methods recognize that a fixed threshold is poorly suited to a growing label space, so they make the threshold adaptive. TACLE [27] introduces a task-adaptive threshold, explicitly motivated by the observation that average prediction confidence decays as more tasks accumulate. Recent work by Wang et al. (2025) [52] also aims for stability: it combines a temporal consistency loss with unlabeled replay, generates pseudo-labels from a frozen past model to regularize new-task learning, and aligns predictions across weakly- and strongly-augmented views. Cui et al. [7, 8] established a standard benchmark for this setting, with protocols that supplement a few labeled samples with unlabeled streams. Across these methods, however, a common weakness remains. Deciding whether to trust an unlabeled sample still comes down to a hard, per-sample confidence criterion, fixed or adaptive. This treats data points independently and discards the dense structural information available in the unlabeled stream.

4. **Structural and topology-preserving methods.** A more recent line of work emphasizes structure more explicitly. The USP framework [11] takes a divide-and-conquer approach, separating the problem into Unlabeled Learning, Memory Stability, and Learning Plasticity with a feature-space reservation mechanism. Even so, its unlabeled-learning component still relies on a confidence threshold to pick pseudo-labels. A number of distillation methods also preserve relational or topological structure to reduce forgetting. For example, relational knowledge distillation [39], topology-preserving CIL (TPCIL) [46], Persistence Homology Distillation (PsHD) [12], and Dynamic Sub-Graph Distillation (DSGD) [13] enforce consistency among feature relations, or between features and prototypes. MAGIC uses no such distillation term. It uses graph structure for a different purpose: plasticity and the estimation of class statistics. Overall, these earlier methods preserve topology for stability, but they stop short of using the topology of the unlabeled stream to improve plasticity or calibration: a gap we address later.

## 2.3. The structural gap: manifold awareness in incremental learning

Taken together, these lines of work reveal a common, unresolved challenge at the intersection of exemplar-free CIL and semi-supervised learning. While pre-trained models offer a stable foundation [58, 62] and adaptive SSL techniques improve data utilization [27], two distinct gaps remain:

1. **Statistical fragility.** State-of-the-art exemplar-free methods rely on accurate first- and second-order statistics to generate replay data. Under sparse supervision, labeled data alone cannot estimate these statistics reliably, and naive pseudo-labeling injects noise that compounds across tasks, progressively corrupting the stored class representations. The paradigm itself is not misguided: generative replay from class statistics is powerful. What it lacks is a reliable way to estimate those statistics when labels are scarce.
2. **Neglect of geometric structure.** Most SSCIL methods exploit the cluster assumption, which states that decision boundaries should fall in low-density regions [2], but they tend to neglect the complementary manifold-smoothness assumption [1] that points nearby in the latent topology likely share a label. Graph-based semi-supervised learning already makes use of this structure through label propagation in the offline [26, 59], streaming [5], and few-shot transductive [35] settings, yet none of these operates under a class-incremental protocol with catastrophic forgetting, and bringing propagation into class-incremental learning remains unexplored: the closest continual method, NNCSL [28], exploits one-hop neighborhood similarity for representation learning rather than manifold diffusion for pseudo-supervision and statistics calibration. As noted above, existing structural methods like PsHD [12] preserve topology for stability, but they do not leverage the topology of unlabeled data to also improve plasticity and statistical calibration.

The two gaps are independent, and patching them requires separate fixes, not a single one. Previous methods tend to choose one side: they either preserve the structure to maintain stability, or threshold on confidence to allow plasticity. Neither approach draws on the geometry of the unlabeled stream to do both at once. We close the structural gap by building a nearest-neighbor graph in feature space and propagating labels across it. In effect, this turns the unlabeled stream from a collection of isolated points into a structured signal, one that can recruit the dense, low-confidence samples that a confidence threshold would simply discard. For the statistical fragility, we use the soft assignments produced by that graph to weight and stabilize the estimation of class statistics. This allows robust generative replay even when supervision is minimal. Importantly, MAGIC does not throw away the distribution-calibration paradigm; it reinforces it. But it does so by replacing confidence thresholding entirely with an alternative that is aware of the data manifold.

# 3. Preliminaries

### 3.1. Problem formulation

We consider a learning system that encounters a sequence of $T$ tasks. At each incremental stage $t \in \{1, 2, \dots, T\}$, the model is presented with a set of new classes $\mathcal{C}_t$, which are disjoint from all previously seen classes (i.e., $C_s \cap C_t = \emptyset$ $for\ all\ s \neq t$). The training data at stage t consists of two distinct components:

- Labeled set $\mathcal{D}_{l,t}$: a small set of labeled samples $\{(x_i, y_i)\}_{i=1}^{N_{l,t}}$ with $y_i \in \mathcal{C}_t$.
- Unlabeled set $\mathcal{D}_{u,t}$: Following the standard semi-supervised CIL protocol, the unlabeled samples at stage $t$ originate from the current classes $C_t$ but lack ground-truth labels.

The objective is to learn a mapping $f_\theta : X \to Y$ that minimizes prediction error over the cumulative label space $C_{1:t} = \bigcup_{s=1}^{t} C_s$, subject to two strict constraints:

- **Zero exemplar storage.** Access to raw data $\mathcal{D}_{1:t-1}$ is strictly forbidden.
- **Label scarcity.** The model must generalize effectively using minimal supervision, relying on the intrinsic manifold structure of the unlabeled data.

### 3.2. Graph-based label propagation

Label Propagation (LP) is a semi-supervised learning algorithm [48, 59, 66] that infers the labels of unlabeled data by diffusing label information through a graph constructed from the data points. Let $\mathcal{X} = \{x_1, \dots, x_l, x_{l+1}, \dots, x_n\}$ be a set of $n$ data points, where the first $l$ points carry labels $y_i \in \{1, \dots, C\}$ and the remaining $u = n - l$ points are unlabeled. The core assumption is the smoothness: points close in the feature space are likely to share the same label. The process has two steps:

**Step 1: graph construction and sparsification.** The data are represented as a graph $\mathcal{G} = (\mathcal{V}, \mathcal{E})$, where the nodes $\mathcal{V}$ correspond to the data points $\mathcal{X}$. The edges $\mathcal{E}$ represent the similarities between these points. An affinity matrix $W \in \mathbb{R}^{n \times n}$ is constructed to quantify these relationships, where $W_{ij}$ reflects the similarity between $x_i$ and $x_j$. While various similarity measures can be employed (e.g., Gaussian RBF kernel, cosine similarity), a general definition often utilizes a kernel function $K(.,.)$:

$$W_{ij} = K(x_i, x_j) \quad if\ i \neq j,\ \ W_{ii} = 0 \tag{1}$$

For efficiency and robustness to noise the graph is sparsified, typically by retaining only the edges to each node's k nearest neighbors.

**Step 2: label diffusion:** The propagation of labels is modeled as a diffusion process on the graph. A probabilistic transition matrix $P$ is defined by row-normalizing the affinity matrix $W$:

$$P_{ij} = \frac{W_{ij}}{\sum_k W_{ik}} \tag{2}$$

where $P_{ij}$ represents the probability of a label transitioning from node $i$ to node $j$. Let $Y^{(m)} \in \mathbb{R}^{n \times C}$ be the soft label matrix at iteration $m$. The algorithm iteratively updates the label distributions of all nodes by aggregating the information from their neighbors:

$$Y^{(m+1)} = \alpha P Y^{(m)} + (1 - \alpha) Y^{(0)} \tag{3}$$

Here, $Y^{(0)}$ represents the initial label matrix where rows corresponding to labeled data are one-hot encoded ground truth vectors, and rows for unlabeled data are initialized to zero (or uniform distributions). The propagation coefficient $\alpha \in (0,1)$, with $(1 - \alpha)$ controlling the clamping to the initial labels, controls the trade-off between the information propagated from neighbors and the initial label information.

The sequence $\{Y^{(m)}\}$ converges to a stationary state $Y^{\infty}$. Row-normalizing this limit yields a soft label distribution over the $C$ classes for each node:

$$y_{soft}(x_i) = \frac{Y_i^{\infty}}{\sum_c Y_{i,c}^{\infty}}, \qquad \hat{y}_i = \arg \max_{c \in \{1,\ldots,C\}} Y_{i,c}^{\infty} \tag{4}$$

In this work we deliberately retain the full soft distribution $y_{soft}$ rather than collapsing it to its argmax: the propagated probabilities act both as continuous training targets (Section 4.3) and as per-sample calibration weights for statistics estimation (Section 4.4).

### 3.3. Class-conditional feature replay

Statistics-based replay methods satisfy the exemplar-free constraint by storing a compact probabilistic model of each class's feature distribution rather than raw samples [19, 58, 64, 65]. Given a fixed feature extractor $\Phi$, the features $\{\Phi(x) : y = c\}$ of a class $c$ are summarized by a class-conditional Gaussian $N(\mu_c, \Sigma_c)$ in $\mathbb{R}^d$, with class mean $\mu_c$ and covariance $\Sigma_c$. To rehearse old classes at stage $t$ without accessing $D_{1:t-1}$, synthetic phantom features are drawn $z \sim N(\mu_c, \Sigma_c)$ for all $c \in C_{1:t-1}$ and replayed alongside the current data.

The validity of this scheme hinges on the stationarity of $\Phi$. If the backbone drifts across tasks, the statistics estimated when a class was first learned no longer describe that class in the current feature space, causing replay to inject stale samples. A frozen $\Phi$ removes this drift by construction and keeps every stored Gaussian valid for the lifetime of the task stream, a property that MAGIC exploits in Section 4.4. For efficiency and stability under sparse supervision we adopt a diagonal covariance, $\Sigma_c = diag(\sigma_c^2)$, so that phantoms are sampled coordinate-wise as $z = \mu_c + \sigma_c \odot \varepsilon, \varepsilon \sim N(0, I)$.

The central difficulty this paper addresses is not the replay mechanism itself but the estimation of $(\mu_c, \sigma_c^2)$ when only a small fraction of the data carrying class $c$ is labeled. Section 4.4 introduces a soft-weighted estimator (SWGC) designed for exactly this scenario.

## 4. The MAGIC framework

MAGIC tackles label scarcity and catastrophic forgetting with two components on top of a standard replay model. The first is threshold-free graph label propagation, which diffuses supervision from the stored old-class means, the current-class prototypes and the sparse current labels to the unlabeled data, without a confidence threshold. The second, Soft-Weighted Geometry Calibration (SWGC), estimates the mean and variance of every new class in the frozen feature space from all images of the task, each weighted by its propagated confidence. Phantom features drawn from these statistics replay the old classes in later tasks.

The replay model is a standard exemplar-free chassis: a frozen backbone, a residual adapter, a cosine classifier and Gaussian latent replay. Keeping it fixed isolates our contribution to two changes: confidence-based pseudo-labeling is replaced by threshold-free graph propagation, and the class statistics are weighted by the propagated soft assignments.

### 4.1. Architectural overview and initialization

MAGIC handles the stability-plasticity trade-off through its architecture. A frozen feature extractor provides stability, while a lightweight plastic head supplies the plasticity needed to absorb new classes. We follow the standard exemplar-free CIL setup: a frozen backbone pre-trained on ImageNet with a cosine-normalized classification head, as used in pre-trained-model CIL methods such as SimpleCIL and APER [62]; FeTrIL [41] likewise keeps its feature extractor frozen after the first task. The cosine classifier itself was introduced for incremental learning by Hou et al. (2019) [24]. Plasticity is confined to a zero-initialized residual adapter, along the lines of parameter-efficient adaptation schemes such as SSF [33] and adapters [25]. Because all learning stays inside this small module, the representation remains stationary. SWGC (Section 4.4) relies on that stationarity to estimate class statistics without drift.

**Frozen backbone Φ.** A ResNet-18 [21] initialized with ImageNet [10] weights, with all parameters and Batch Normalization statistics permanently frozen. This yields a stationary coordinate system $Z_{frozen} \subset \mathbb{R}^d$ in which the mapping $z = \Phi(x)$ is constant for the lifetime of the task stream, eliminating the feature drift that undermines statistics-based replay in fully plastic backbones.

**Plastic head.** All adaptation is carried by a residual adapter followed by a cosine classifier. For a frozen feature $z = \Phi(x)$, the adapter computes

$$h_\theta(z) = z + LayerNorm(W\, z + b), \tag{5}$$

a single affine layer with feature normalization applied through a skip connection, and we call $h_\theta(z)$ the adapted feature. The parameters $W$ and $b$ are zero-initialized, so at the start of training $LayerNorm(W\, z + b) = 0$ and $h_\theta$ is exactly the identity map $z \mapsto z$, departing from the frozen representation only as far as the incremental data require. The initialization is done once, before the first task; afterwards the adapter keeps its parameters from task to task and is never reset. This design anchors early-task optimization to the robust frozen features and avoids the instability that a freely initialized adapter would introduce.

**Cosine classifier.** A plain linear layer computes the score for class $c$ by the dot product $w_c^T h_\theta(z)$. This score is sensitive to the norms of both the feature vector and the weight vector. In class-incremental learning, this norm sensitivity induces magnitude bias: recently learned classes tend to acquire larger weight norms and are favored independently of their semantic relevance. We therefore score classes by cosine similarity on a hyperspherical manifold,

$$logit_c = s.\frac{h_\theta(\Phi(x))^T w_c}{\|h_\theta(\Phi(x))\|\|w_c\|} \tag{6}$$

where $w_c$ is the learnable weight vector of class $c$ and $s = 30$ is a fixed scale that sharpens the softmax and stabilizes convergence. Normalizing both features and weight vectors gives every class equal voting power regardless of when it was learned. For brevity we write $L_{CE}(h_\theta(z), y)$ for the cross-entropy of these cosine logits against a label $y$, and use this convention for all latent-space losses in Section 4.3. The classifier holds a weight vector $w_c$ for every class of the stream from the start; the logits of classes not yet seen are masked, so every softmax in MAGIC runs over the classes seen so far.

**Latent Gaussian Manager $\mathcal{M}$.** Instead of storing raw exemplars, we maintain a lightweight manager $\mathcal{M}$ that stores, for every seen class $c$, a diagonal-covariance class-conditional Gaussian in the frozen feature space: a mean $\mu_c \in \mathbb{R}^d$ and a per-dimension variance $\sigma_c^2 \in \mathbb{R}^d$. We adopt a diagonal covariance, rather than a full one, for memory efficiency and stability under sparse supervision. Phantom features for replay are then drawn as $z = \mu_c + \sigma_c \odot \varepsilon$ with $\varepsilon \sim \mathcal{N}(0, I)$.

### 4.2. Graph construction and label propagation

To exploit unlabeled data we build a $k$-nearest-neighbor graph in feature space and propagate the sparse labels transductively. Graph-based label propagation is a classical semi-supervised technique [26, 59]. We do not claim novelty in the propagation mechanism itself. Instead, we use it in two ways that are new to the class-incremental setting: (i) as a threshold-free alternative to confidence-based pseudo-labeling, and (ii) to couple its soft outputs with statistical calibration through SWGC (Section 4.4). In the spirit of the per-example certainty weights of Iscen et al. [26], we weight each propagated pseudo-label by its squared graph confidence.

Before training, the framework constructs the propagation graph that propagates labels from sparse labeled anchors to the abundant unlabeled data. Within a task we drop the task index and write $D_L$ and $D_U$ for its labeled and unlabeled sets $D_{l,t}$ and $D_{u,t}$.

To balance the historical and current class distributions, the graph nodes $\mathcal{V}$ are assembled from four sources:

- **Old-class prototypes.** The stored means $\mu_c \in \mathbb{R}^d$ for all $c \in \mathcal{C}_{old}$ are retrieved from the Latent Gaussian Manager $\mathcal{M}$. These serve as stationary, high-confidence anchor nodes for previous knowledge.
- **Current-class prototypes.** For every current class $c \in \mathcal{C}_t$, the mean of the frozen features of its labeled images.

- **Current real data.** Features extracted by the frozen backbone $\Phi$ for all current real samples: $\mathcal{Z}_{curr} = \{\Phi(x_i) | x_i \in \mathcal{D}_L \cup \mathcal{D}_U\}$.
- **Current augmented anchors.** To strengthen the few labeled seeds, the labeled samples of the current task are replicated and jittered with Gaussian noise $\epsilon \sim \mathcal{N}(0, \sigma^2(z)I)$, where the per-sample scale $\sigma(z) = 0.1 \parallel z \parallel / \sqrt{d}$ is set relative to the feature norm, and the jittered copies are L2-normalized. Each current class receives $\max(20, \min(100, \lfloor N_t/(5|\mathcal{C}_t|) \rfloor))$ copies, where $N_t$ is the number of images of task $t$: 100 per class on CIFAR-100, 20 on CUB-200 and about 24 on ImageNet-R. These copies form dense clusters that anchor the current classes in the latent space; removing them costs up to 3.6 AIA points at 1% labels (Section 5.5).

Phantom samples are not used as graph nodes; each old class enters the graph only through its stored mean. Because these means are seeds, the propagated soft labels span all classes seen so far, and an image that resembles an old class keeps part of its label mass on that class. Under the protocol of Section 3.1 every unlabeled image belongs to a current class, so the old-class means do not serve to label old-class images; their measured effect is on accuracy and forgetting. Removing all prototype nodes (old and current) lowers AIA on CIFAR-100 by about 3 points and raises its forgetting by 13–18 points, but raises AIA slightly on CUB-200 and ImageNet-R at 1% labels (Section 5.5). A side effect is that the propagated confidences are spread over more classes as tasks accumulate, so the weights of Eq. (8) shrink along the task sequence: at 1% labels, the mean weight of an unlabeled image falls from 0.50 in the first task to 0.18 in the last on CIFAR-100, and from 0.40 to 0.06 on CUB-200.

The adjacency matrix $W$ is constructed using $k$ nearest neighbors under cosine similarity:

$$W_{ij} = \begin{cases} \exp\left(\cos(z_i, z_j) / T_{prop}\right) & if\ z_j \in kNN(z_i) \\ 0 & otherwise, \end{cases} \tag{7}$$

where $T_{prop}$ is the propagation temperature. Unlike in Eq. (1), each node is counted among its own $k$ nearest neighbors, so the graph contains self-loops; the adjacency is then symmetrized as $(W + W^{\top})/2$. Label information is then diffused from the seed nodes (labeled samples, prototypes and augmented anchors) by the procedure of Section 3.2, with $\alpha = 0.8$ and 50 iterations, producing soft targets $Y_{soft}$ over the seen classes for every image of the task (one row $y_{soft}(x)$ per image $x$); those of $\mathcal{D}_U$ serve as pseudo-supervision during training.

### 4.3. Training objective

The main training phase optimizes the plastic head $h_\theta$ (adapter and classifier) with a composite objective of three terms:

**Supervised and semi-supervised losses** ($\mathcal{L}_{Sup}$, $\mathcal{L}_{SSL}$)**:** Each training step uses a batch of 16 labeled and 112 unlabeled images of the current task, at every label ratio. The ratio changes the pools that these images come from, not the batch: on CIFAR-100 a task has 50 labeled images at $r = 1\%$ and 500 at $r = 10\%$. An epoch is one pass over the unlabeled images; the labeled images are reshuffled and reused whenever they run out, so on CIFAR-100 at $r = 1\%$ each labeled image appears in about 14 batches per epoch. The only augmentation is a random horizontal flip. $\mathcal{L}_{Sup}$ is the cross-entropy on the labeled images. Every unlabeled image of the batch contributes to $L_{SSL}$, weighted continuously by its squared graph confidence. There is no confidence gate: rather than accept or reject a sample by a threshold, we let each sample's contribution scale with its graph-derived reliability, so structurally reliable low-confidence samples are down-weighted rather than discarded.

$$\mathcal{L}_{SSL} = -\frac{1}{|\mathcal{B}_U|}\sum_{x\in\mathcal{B}_U} w_x \cdot \sum_c Y_{soft}(x,c). \log\ p(c|x), \qquad w_x = \left(\max_c Y_{soft}(x,c)\right)^2 \tag{8}$$

where $B_U$ denotes the unlabeled images of the batch, and the softmax $p(\cdot\ |x)$ is taken over the seen classes. Squaring sharpens the down-weighting of ambiguous propagations while retaining structurally reliable low-confidence samples rather than discarding them by a threshold.

**Replay ($\mathcal{L}_{Replay}$).** To replay old knowledge we sample phantom features $z \sim \mathcal{N}\left(\mu_c, \text{diag}(\sigma_c^2)\right)$ from $\mathcal{M}$ in the frozen feature space, for every old class $c$. At every step each old class receives $n_t = \min\left(32, \max(1, \lfloor 128/|\mathcal{C}_{old}|\rfloor)\right)$ phantoms: at most 128 in total while there are at most 64 old classes, and one per old class beyond that. The phantoms pass through the adapter and the classifier like real features, and $\mathcal{B}_{ph}$ denotes the phantoms of a batch:

$$\mathcal{L}_{Replay} = \frac{1}{|\mathcal{B}_{ph}|}\sum_{(z,y)\in\mathcal{B}_{ph}} \mathcal{L}_{CE}\left(h_\theta(z), y\right) \tag{9}$$

The final objective is the weighted sum of the three terms:

$$\mathcal{L}_{Total} = \mathcal{L}_{Sup} + \mathcal{L}_{SSL} + \lambda_{Replay}\mathcal{L}_{Replay} \tag{10}$$

### 4.4. Soft-Weighted Geometry Calibration (SWGC)

For exemplar-free rehearsal, we model each old class as a class-conditional Gaussian with diagonal covariance in feature space and replay synthetic "phantom" features sampled from these statistics. This strategy is well established by prototype- and statistics-based replay methods such as IL2A [64], SLCA [58], and FeCAM [19]. Our departure lies not in the replay mechanism itself, but in how the statistics are estimated under sparse supervision, through SWGC.

At the end of each task the raw data are discarded, so the statistics $\left(\mu_c, \sigma_c^2\right)$ of the new classes must be estimated and stored in $M$ first. With one to five labeled images per class, the labeled images alone give poor estimates, and a confidence threshold adds few pseudo-labeled images to them (Section 5.4). Soft-Weighted Geometry Calibration (SWGC) therefore uses every image of the task, weighted by its propagated soft assignment, to estimate the class statistics in the frozen feature space.

A second graph is first built on the adapted features $h_\theta(z)$ of the task images, produced by the trained head, which separate the classes better than the frozen features. This graph contains only the images of the task, without prototypes or anchors, and only the labeled images are seeds; it uses the same $k$, $T_{prop}$ and $\alpha$ as the first graph. Label propagation on it gives soft labels for all images of the task, which weight the mean and variance of each new class:

$$\mu_c = \frac{\sum_i w_{i,c}^2 z_i}{\sum_i w_{i,c}^2}, \quad \sigma_c^2 = \frac{\sum_i w_{i,c}^2 (z_i - \mu_c)^2}{\sum_i w_{i,c}^2} \tag{11}$$

where $w_{i,c}$ is the soft assignment of image $i$ to class $c$ given by this second propagation. For an image with a ground-truth label of class $c$ we set $w_{i,c} = 1$, which anchors the estimate to verified data; all other weights are the propagated soft assignments. Statistics are computed on the frozen features, so they remain valid for the rest of the stream, while the weights come from the adapted features. Variances are floored at $10^{-5}$.

These statistics are stored in the Latent Gaussian Manager $\mathcal{M}$, allowing the system to safely discard raw data and proceed to Task $t + 1$.

### 4.5. Algorithmic summary

The complete training procedure for MAGIC is summarized in Algorithm 1, and Figure 1 shows the pipeline. The process runs sequentially over tasks $\mathcal{T}_1, \dots, \mathcal{T}_T$.

**Algorithm 1:** MAGIC SSCIL Training Framework

**Input:** Task stream $\mathcal{T}_1, \dots, \mathcal{T}_T$, frozen backbone $\Phi$, plastic head $h_\theta$
**Output:** optimized model, Latent Gaussian Manager $\mathcal{M}$

**for** $t = 1$ **to** $T$ **do**
  *// **Phase 1: Graph pseudo-labels***
  extract $Z \leftarrow \Phi(D_L \cup D_U)$; read the old-class means $\mu$ from $\mathcal{M}$
  build the $k$-NN graph over $\mu \cup Z \cup$ current prototypes $\cup$ anchors
  $Y_{soft} \leftarrow$ threshold-free label propagation (Section 3.2)
  *// **Phase 2: Training***
  **for** *each iteration* **do**
    draw labeled and unlabeled images and phantoms from $\mathcal{M}$
    minimize $\mathcal{L}_{Total} = \mathcal{L}_{Sup} + \mathcal{L}_{SSL} + \lambda_{Replay}\mathcal{L}_{Replay}$
  **end for**
  *// **Phase 3: Soft-Weighted Geometry Calibration***
  build a second graph on $h_\theta(Z)$, seeded by the labeled images
  compute $\mu_c$, $\sigma_c^2$ on the frozen features with weights $w_{i,c}^2$
  store the statistics in $\mathcal{M}$; discard all raw data
**end for**

**Phase 1: graph pseudo-labels.** For each task, the labeled set $\mathcal{D}_L$ and the unlabeled set $\mathcal{D}_U$ are encoded by the frozen backbone, and a $k$-NN graph is built on these features together with the old-class means stored in $\mathcal{M}$, the current-class prototypes and jittered copies of the labeled images; phantoms are not used as nodes. Label propagation yields soft targets $Y_{soft}$ for all images of the task.

**Phase 2: training.** The main loop optimizes the adapter and the classifier with $\mathcal{L}_{Total}$ (Eq. (10)): the supervised and semi-supervised terms drive adaptation to the new classes, and the phantoms replay the old ones.

**Phase 3: Soft-Weighted Geometry Calibration.** On completion of the task, the adapted features of its images are computed with the trained plastic head, the second graph is built on them, and the resulting soft labels weight the mean and variance of the frozen features of each new class. These statistics are stored in $M$, and the raw data are then discarded.

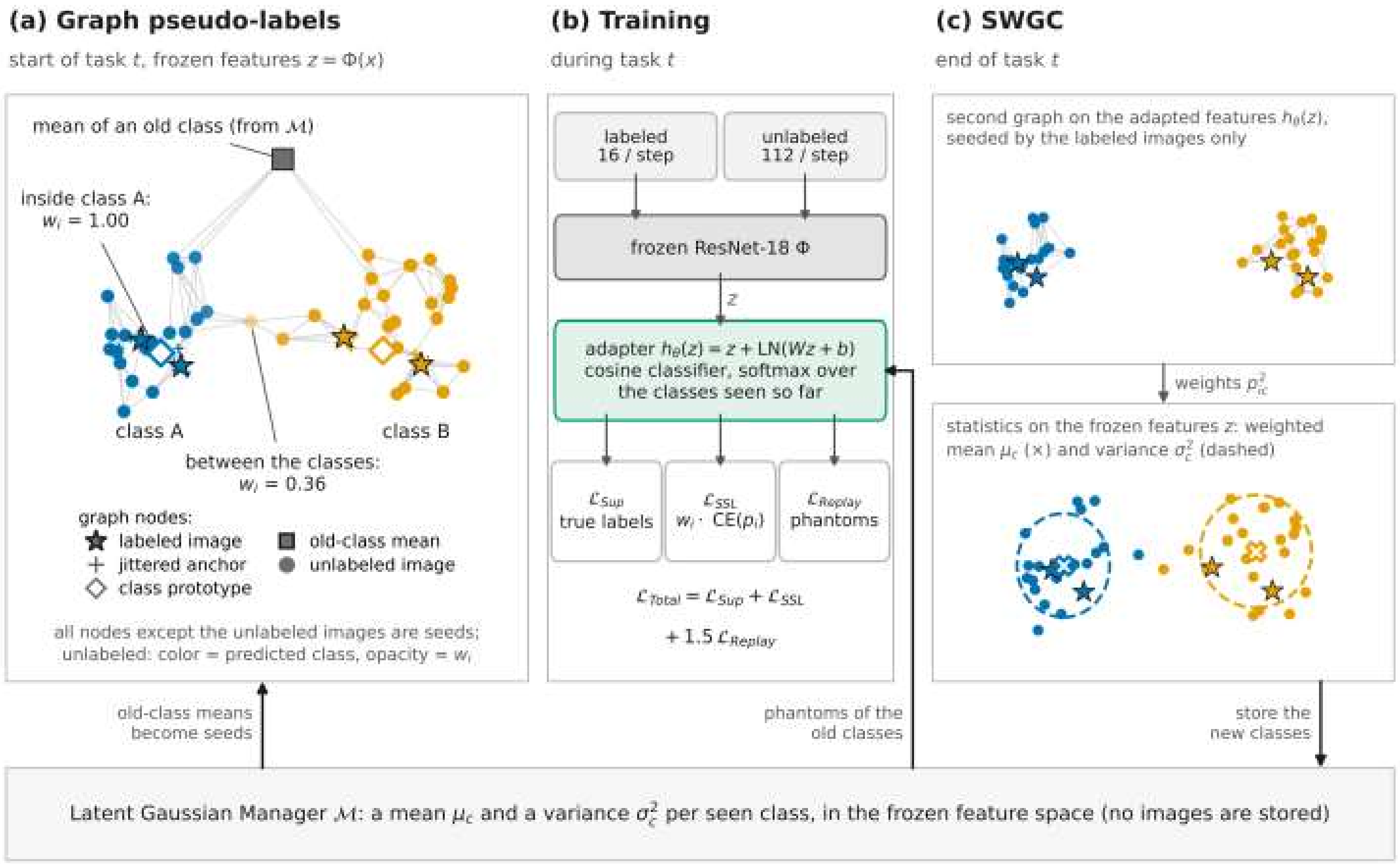

**Figure 1.** *The MAGIC pipeline for one task $t$. (a) At the start of the task, labels propagate over a $k$-NN graph of frozen features whose seeds are the labeled images, their jittered copies, the current-class prototypes and the stored means of the old classes; every image receives a soft label $p_i$ and a weight $w_i = (\max_c p_{ic})^2$ (its row of $Y_{soft}$ and the weight $w_x$ of Eq. (8)), with no threshold. (b) The adapter and the cosine classifier are trained on top of the frozen backbone with the supervised loss, the weighted unlabeled loss and replay of phantom features drawn from the stored Gaussians. (c) At the end of the task, a second propagation over the adapted features gives the weights $p_{ic}^2$ (the $w_{i,c}^2$ of Eq. (11)); SWGC computes the weighted mean and variance of each new class on the frozen features and stores them in $\mathcal{M}$, where they serve later tasks as graph nodes and as the source of phantoms.*

## 5. Experimental study

In this section, we compare MAGIC with eleven baselines (Section 5.2), measure how much each method gains from unlabeled data (Section 5.3), test inside MAGIC which pseudo-labels and which gate produce its gain (Section 5.4), and remove its components one at a time (Section 5.5).

### 5.1. Experimental setup

#### *5.1.1. Datasets and incremental protocol*

We evaluate on three standard benchmarks: CIFAR-100 [30], CUB-200-2011 [49], and ImageNet-R [22]. The three cover complementary scenarios: coarse-grained recognition, fine-grained recognition with very few images per class, and a strong distribution shift relative to the ImageNet pre-training data. Dataset statistics are given in Table 1.

Each dataset is divided into $T = 10$ tasks with disjoint class sets and an equal number of classes per task; we do not use an enlarged first task (which we also call the base session). Classes are presented in the dataset's native index order, which is the same for every method, label ratio and run. Appendix C reports how MAGIC's accuracy changes under other class orders. At task $t$ a method sees only $D_{l,t}$ and $D_{u,t}$, and data from earlier tasks are not revisited. All methods are exemplar-free except USP, which is replay-based by design: it keeps a total buffer of 5,120 exemplars, allocated evenly over the classes seen so far and selected by herding [42, 55].

Images are resized to $224 \times 224$ with bicubic interpolation, the resolution at which the backbone was pre-trained, following common practice for CIL with pre-trained models [61, 62]. At the native $32 \times 32$ resolution of CIFAR-100, the last residual stage of an ImageNet-style ResNet-18 operates on a $1 \times 1$ feature map and the pre-trained BatchNorm statistics no longer match the activations they normalize, so this step is necessary rather than cosmetic.

**Table 1.** *Datasets and incremental protocol. Labeled budgets are per class, computed as* $max(1, \lceil r.n_c \rceil)$. *ImageNet-R has no official split: each class is split 80/20 at random (seed 42). Its classes are imbalanced, so its labeled counts are means over classes (1–4, 2–18 and 4–35 images per class at the three ratios).*

| Dataset | Classes | Train | Test | Tasks × classes | Labeled images / class | | |
|---|---|---|---|---|---|---|---|
| | | | | | **r=1%** | **r=5%** | **r=10%** |
| CIFAR-100 | 100 | 50000 | 10000 | 10×10 | 5 | 25 | 50 |
| CUB-200 | 200 | 5994 | 5794 | 10×20 | 1 | 2 | 3 |
| ImageNet-R | 200 | 23918 | 6082 | 10×20 | 1.7 | 6.5 | 12.4 |

### *5.1.2. Compared methods*

We compare MAGIC against nine methods covering the main families of exemplar-free CIL, plus the two SSCIL methods that address our setting directly. Table 2 lists them along the three axes that matter for this study: whether raw exemplars are stored, whether second-order class statistics are stored, and how the unlabeled pathway is gated.

TACLE [27] and USP [11] are native SSCIL methods, so they keep their published semi-supervised machinery. The other nine baselines were originally designed for fully supervised learning and cannot use unlabeled data on their own. Evaluating them as-is on $r\%$ of the labels would conflate two distinct weaknesses: a weak CIL mechanism and a missing SSL mechanism. To avoid this conflation, we equip each of them with a FixMatch-style pseudo-labeling module [45]. In this module, a weakly augmented view supplies the pseudo-label, a strongly augmented view receives the consistency loss, and the unsupervised loss weight is $\lambda_u = 1$ throughout. We append an FM subscript to the method's name (e.g., $\text{PASS}_{\text{FM}}$) to mark these FixMatch-augmented variants in the tables and figures; in the running text we drop the subscript, and a baseline's name always denotes its semi-supervised variant unless stated otherwise. The exact integration point differs across methods because what gets trained, and when, varies: some methods fine-tune the backbone on every task, some only on the base session, and some never. Table A1 summarizes what each method trains, where the unlabeled signal enters, and the inference rule. Appendix A describes each integration in detail. For LDC, we treat it similarly to the other supervised baselines. We pair its core contribution (the drift-compensation projector) with a backbone trained by the LwF + FixMatch objective of our LwF baseline, rather than the self-supervised backbone (CaSSLe/PFR) from the original work [14, 17] that targets the cold-start setting. Under our warm-start protocol with a pre-trained backbone, LDC therefore consumes unlabeled data through FixMatch on the backbone and through its label-free projector on top.

The threshold policies in Table 2 are worth a remark. All nine integrated baselines use a fixed threshold, as FixMatch prescribes. USP does not discard its low-confidence samples: it splits each unlabeled batch at the threshold and relabels the low-confidence part by nearest class mean. TACLE goes further and schedules its threshold downward across tasks, from 0.90 at the base session toward 0.65, via an inverse-sigmoid rule. That the authors of a native SSCIL method found a decaying threshold necessary suggests that one fixed confidence gate is hard to set for a whole task sequence; Section 5.4 tests fixed and adaptive gates inside MAGIC. MAGIC uses no confidence threshold at any stage. Since the absence of a threshold is central to our claims, all baseline thresholds are kept at their published or best-known values, which makes the comparison conservative with respect to MAGIC.

Finally, our experiments also contain key reference points to contextualize performance. First, for every method at each label ratio, we run a labeled-only variant trained on the labeled subset with the unlabeled stream discarded; this isolates what each method gains from unlabeled data. Second, for the supervised methods, we include a fully supervised variant trained with all labels, which shows how much room for improvement remains when supervision is not the bottleneck.

**Table 2.** *Compared methods.* ***Exem.****: stores raw exemplars.* ***2nd-ord.****: stores second-order statistics (per-class covariances or variances, or a shared Gram matrix).* ***Unlabeled pathway****: native = published by the authors; FixMatch (ours) = added by us (Section 5.1.2 and Appendix A).* $\tau$ ***policy****: how the pathway is gated by confidence (fixed: one threshold for all tasks; split: images below the threshold are relabeled, not discarded; decaying: the threshold falls over the tasks; none: no threshold).*

| Method | Family | Exem. | $2^{nd}$-ord. | Unlabeled pathway | $\tau$ policy |
|---|---|---|---|---|---|
| $\text{LwF}_{\text{FM}}$ | Regularization | × | × | FixMatch (ours) | Fixed |
| $\text{SimpleCIL}_{\text{FM}}$ | Prototype | × | × | FixMatch (ours) | Fixed |
| $\text{APER}_{\text{FM}}$ | Prototype + adapter | × | × | FixMatch (ours) | Fixed |
| $\text{PASS}_{\text{FM}}$ | Prototype + augmentation | × | × | FixMatch (ours) | Fixed |
| $\text{FeTrIL}_{\text{FM}}$ | Feature translation | × | × | FixMatch (ours) | Fixed |
| $\text{FeCAM}_{\text{FM}}$ | Mahalanobis prototype | × | ✓ | FixMatch (ours) | Fixed |
| $\text{RanPAC}_{\text{FM}}$ | Random projection + analytic | × | ✓ | FixMatch (ours) | Fixed |
| $\text{SLCA}_{\text{FM}}$ | Slow learner + alignment | × | ✓ | FixMatch (ours) | Fixed |
| $\text{LDC}_{\text{FM}}$ | Drift compensation | × | × | FixMatch (ours) + projector | Fixed |
| USP (iCaRL-based) | SSCIL, replay | ✓ | × | Native | Fixed, split |
| TACLE | SSCIL, Gaussian | × | ✓ | Native | Decaying |
| **MAGIC (ours)** | SSCIL, geometric | × | ✓ | Native | None |

### *5.1.3. Semi-supervised protocol*

**Label budget.** Within each task we keep a stratified per-class subset of ratio $r \in \{1\%, 5\%, 10\%\}$ of the training images as the labeled set $D_{l,t}$; the rest forms the unlabeled set $D_{u,t}$ and its labels are discarded. The per-class count is $max(1, \lceil r.n_c \rceil)$, so every class has at least one labeled image even at the smallest budget. At $r = 1\%$ this gives 5 labeled images per class on CIFAR-100, and a single image per class on CUB-200. ImageNet-R is class-imbalanced: it has 1 to 4 labeled images per class, 1.7 on average (Table 1). The labeled subset is not fixed: every run draws its own subset with its own seed, so the spread over seeds that we report includes the effect of which images are labeled. One deviation favors three baselines: their scripts compute the labeled fraction as one minus the unlabeled fraction, and the floating-point rounding gives $\mathrm{LwF_{FM}}$, $\mathrm{PASS_{FM}}$ and $\mathrm{FeTrIL_{FM}}$ one extra labeled image per class on CIFAR-100 at $r = 1\%$ and 5% (6 and 26 instead of 5 and 25), and on 3 and 14 classes of ImageNet-R.

**Unlabeled stream.** $D_{u,t}$ contains images of the classes of task $t$ only, mirroring the class-incremental partition of the labeled data, and is not carried across tasks. The pseudo-labeling baselines therefore assign pseudo-labels over the current task's classes only, implemented by masking or slicing the logits of previously seen classes, and the native SSCIL baselines make the same assumption. MAGIC differs here: because the stored old-class means are seeds of its graph (Section 4.2), its soft labels and its unlabeled loss span all classes seen so far.

**Batch composition**. Every method uses a total batch of 128 real images (with the one exception given below), and its split into labeled and unlabeled images is the same at every label ratio: the ratio changes how many labeled images a task has (Table 1), so with few labels the same labeled images recur in many batches. The FixMatch-integrated baselines split it as 16 labeled + 112 unlabeled ($\mu = 7$), following Sohn et al. [45], with one exception: RanPAC performs no gradient training, so labeled and unlabeled images are processed in separate feature-extraction passes at batch 128. USP and TACLE keep their native 64+64 split, and MAGIC also uses 16+112 and adds its phantom features for replay (Section 4.3). On CUB-200 at $r = 1\%$, where a task holds only 20 labeled images, we set USP to 32+32.

**Confidence thresholds**. Native methods keep their published values. USP uses 0.95 on CIFAR-100, 0.50 on CUB-200, and 0.90 on ImageNet-R. TACLE uses the inverse-sigmoid schedule described above ($\alpha = 0.5, \beta = 0.65$). For the integrated baselines we set one threshold per dataset, shared across methods: 0.95 on CIFAR-100, 0.85 on CUB-200 and 0.90 on ImageNet-R.

One caveat applies to any comparison of thresholds across methods: the probability compared against $\tau$ is not computed identically everywhere. FeCAM applies a softmax to cosine similarities divided by 0.05, and the other integrated baselines use plain softmax probabilities. The same nominal $\tau = 0.95$ therefore admits different fractions of the unlabeled batch in different methods. We report each method's confidence parameterization in Table A2 rather than forcing a common one, since these settings are part of the published methods; the incommensurability of nominally equal thresholds is itself an instance of the calibration problem this paper is about.

#### *5.1.4. Implementation details*

**Backbone.** All methods use the same ResNet-18 [21] initialized from the torchvision [40] IMAGENET1K_V1 checkpoint. No method sees additional pre-training data.

**Shared settings and native settings.** We hold fixed everything that defines the experimental condition: the backbone and its initialization, input resolution, class order, task partition, the label-split rule and per-class counts, the total batch size of 128, the unlabeled stream, and the evaluation protocol (the two small deviations are stated in Section 5.1.3). We do not force a single optimization recipe on all methods. A published method's optimizer, learning rate, schedule and weight decay are settings its other components were tuned around. Overriding them penalizes the method for something unrelated to its CIL mechanism. A concrete example is PASS, which is published with Adam: training it with the SGD recipe used elsewhere in our study lowers its AIA on CIFAR-100 at $r = 10\%$ from 56.81 to 49.41. Each method therefore runs with the recipe from its paper and official code, listed in full in Table A2. For every method whose native optimizer differs from SGD we also ran the common SGD configuration and kept whichever was stronger; in all cases this was the native recipe. Note that the baselines receive their per-dataset published settings, while MAGIC uses one configuration for all datasets and label ratios.

**Training budget.** Gradient-trained methods use 20 epochs per task on CIFAR-100, 30 on ImageNet-R and 40 on CUB-200. TACLE splits its budget into 14 representation epochs and 6 classifier-alignment epochs on CIFAR-100. APER and FeCAM train only in the base session: they use the same budget there and run their prescribed closed-form or single-pass computation on later tasks. FeTrIL trains a linear head on frozen features at every task, and RanPAC involves no gradient training at all, like SimpleCIL in its original form. Realized budgets are listed per method in Table A2.

**Backbone adaptation**. LwF, PASS, SLCA, TACLE, LDC and USP fine-tune the full backbone on every task, with the backbone learning rate reduced relative to the head ($100 \times$ lower for LwF, SLCA, TACLE, LDC and FeCAM's base session; $10 \times$ lower for PASS and USP). APER trains only its SSF parameters and classifier in the base session and freezes everything afterwards. FeTrIL and RanPAC keep the pre-trained backbone frozen throughout, and SimpleCIL trains only its cosine head. MAGIC also keeps the backbone frozen and trains a residual adapter and cosine classifier on top of the frozen features.

**Augmentation**. For the baselines, the weak view is a random crop with horizontal flip; the strong view is RandAugment [6]. PASS is the exception: RandAugment's geometric operations, rotation in particular, interfere with the rotation-prediction task that PASS's self-supervised loss is built on, so for PASS we use a rotation-safe strong policy of comparable strength (color jitter, random grayscale, Gaussian blur, random erasing).

**Normalization under a matched backbone**. Four methods (APER, RanPAC, SLCA, TACLE) were introduced on ViT backbones, whose LayerNorm keeps no running statistics. Under ResNet-18, BatchNorm running statistics drift as new tasks are learned, at a rate essentially independent of the learning rate. This is harmful only for methods that keep adapting the backbone and reuse class statistics estimated in an earlier feature space: TACLE (stage-2 class means and covariances) and SLCA (per-class Gaussians). For these two we keep BatchNorm layers in eval mode during training while leaving all weights trainable, which restores the consistency between the space in which statistics are stored and the space in which inference runs, a consistency their original LayerNorm backbones provide implicitly. SLCA's slow learning rate does not solve the problem on its own, because BatchNorm statistics drift regardless of the weight learning rate. APER and RanPAC freeze the backbone after (or before) the base session and are unaffected, as are FeTrIL, FeCAM and SimpleCIL. PASS counteracts representation drift with its own machinery (feature distillation and prototype augmentation) and keeps standard BatchNorm behavior, as does LwF, which stores no statistics at all.

**MAGIC configuration.** The frozen backbone feeds a residual adapter (a linear layer with LayerNorm, zero-initialized once before the first task) and a cosine classifier with scale $s = 30$, trained with Adam (learning rate $10^{-4}$, weight decay $10^{-5}$) and a new optimizer at each task; the learning rate is constant, and the number of epochs per task follows the training budget above. We set $\lambda_{Replay} = 1.5$; $\mathcal{L}_{Sup}$ and $\mathcal{L}_{SSL}$ are unweighted. Both graphs use $k = 25$ neighbors (counting the node itself), temperature $T_{prop} = 0.2$, $\alpha = 0.8$ and 50 propagation iterations. Each step replays $\min\big(32, \max(1, \lfloor 128/|\mathcal{C}_{old}|\rfloor)\big)$ phantoms per old class. Because the backbone is frozen and the only augmentation is a random horizontal flip, the features of every image and of its mirror image are computed once and cached. Apart from the number of epochs, all values are the same for every dataset and label ratio, with no per-dataset or per-ratio tuning.

**Reproducibility.** All methods are implemented in PyTorch. The experiments were run on an NVIDIA GeForce RTX 4090. Every baseline configuration is repeated with three seeds (42, 127, 2026); the seed also chooses the labeled images (Section 5.1.3). MAGIC trains only an adapter and a classifier on cached features, so one run takes less than two minutes on a T4 GPU, and we repeat every MAGIC configuration with five seeds: the same three and two more (7 and 99). The two extra seeds serve the paired comparisons inside MAGIC (Sections 5.4 and 5.5). With the baselines' three seeds alone, MAGIC's AIA and last-task accuracy change by at most 0.4 points, and its comparison with the baselines changes in one of 18 cases (Appendix C). We report the mean and the sample standard deviation over seeds.

#### *5.1.5. Evaluation metrics*

Let $a_{t,i}$ denote the top-1 accuracy on the test set of task $i$ after training on task $t$ ($i \leq t$), and $A_t$ the top-1 accuracy on the pooled test set $\bigcup_{i\leq t}\mathcal{D}_i^{\text{test}}$ of all classes seen so far. We report the average incremental accuracy (AIA) [42], the accuracy after the final task ($A_T$), and the average forgetting $F_T$ [3]:

$$AIA = \frac{1}{T}\sum_{t=1}^{T} A_t \tag{12}$$

$$F_T = \frac{1}{T-1}\sum_{i=1}^{T-1} \max\left(0, \max_{t\in\{i,\dots,T-1\}} a_{t,i} - a_{T,i}\right) \tag{13}$$

Higher AIA and $A_T$ are better; lower $F_T$ is better. The clamp in Eq. (13) means that a task whose accuracy improves over the sequence contributes zero forgetting rather than a negative value.

$A_t$ is micro-averaged over the pooled test set. Task test sets differ in size and ImageNet-R is class-imbalanced, so micro- and macro-averaging do not coincide, and the public implementations of the compared methods are split between the two conventions. We harmonized all implementations to the micro-averaged convention of Eq. (12). All reported numbers are therefore directly comparable. Each method is evaluated with its own native classifier rather than a common nearest-mean rule, since the comparison is between complete systems and several methods' classifiers are part of their contribution. For LDC this is the nearest-class-mean rule over its drift-compensated prototypes, not the auxiliary linear head its implementation also maintains.

#### *5.1.6. Statistical protocol*

MAGIC and the baselines were run on different random labeled subsets, one per seed, so we compare them setting by setting with Welch's unequal-variance t-test [54] over the runs (five for MAGIC, three for each baseline) and report the difference with its 95% confidence interval. The test allows unequal numbers of runs; Appendix C repeats it with MAGIC restricted to the baselines' three seeds. We test the nine settings separately and do not pool them into one test, because the three label ratios of a dataset share its images and are not independent. Inside MAGIC (Sections 5.4 and 5.5), every variant is run with the same five seeds as the full model, and therefore with the same labeled images and class order; there we report the mean paired difference with its 95% t-interval. An interval that contains zero means that the data cannot separate the two methods; it does not show that they are equal. The intervals are not adjusted for multiple comparisons.

### 5.2. Main comparison

Tables 3–5 report average incremental accuracy (AIA), last-task accuracy $A_T$ and average forgetting $F_T$ for all twelve methods on the three datasets and three label ratios, and Table 6 compares MAGIC with every baseline in each setting.

**With 1% of the labels, MAGIC is the best method on CUB-200, level with the best on ImageNet-R and second on CIFAR-100.** It reaches an AIA of 63.59 on CIFAR-100, 37.32 on CUB-200 and 19.78 on ImageNet-R. On CIFAR-100 the strongest baseline is TACLE (65.76): MAGIC is 2.2 points below it (95% interval −3.8 to −0.5), level with SLCA (63.36) and FeTrIL (59.69), and ahead of the other eight baselines. On ImageNet-R the strongest baseline is SLCA (20.14); the difference, −0.4 points, lies well inside its 95% interval (−1.1 to +0.4), so the data

cannot separate the two methods, and MAGIC is ahead of nine of the eleven baselines (Table 6). On CUB-200, MAGIC leads the strongest baseline, APER (28.41), by 8.9 points (interval +2.7 to +15.2), and each of its five runs scores higher than each APER run. In last-task accuracy MAGIC is ahead at this budget on CUB-200 and ImageNet-R, by 10.0 points over APER and by 1.5 over SLCA; on CIFAR-100 it is 3.6 points above SLCA (54.38 vs. 50.76) and 1.8 below TACLE (56.20), a difference whose 95% interval includes zero (−3.8 to +0.2; Tables 4 and 6).

**On fine-grained CUB-200, MAGIC has the highest mean AIA at every label ratio** (37.32, 45.19 and 47.77), but its lead over APER shrinks as labels are added: 8.9 points at 1%, 4.9 at 5% (interval +1.4 to +8.4) and 2.0 at 10%, where the interval (−1.9 to +5.9) includes zero. In last-task accuracy the lead holds at every ratio (+10.0, +7.0 and +4.5 points; Table 6). A CUB-200 class has about 30 training images, so at these ratios it has one to three labeled images, and its statistics must come mostly from unlabeled images. This is where MAGIC gains most from unlabeled data (+14.8, +13.5 and +10.0 AIA points; Section 5.3), and where its stored class means are two to three times closer to the true means than means computed from the labeled images alone (Section 5.4).

**At 5% and 10% labels on CIFAR-100 and ImageNet-R, methods that adapt the backbone or model full covariances overtake MAGIC.** On CIFAR-100, SLCA leads by 5.1 points at 5% and by 6.3 at 10% (73.14 and 75.29 vs. 68.05 and 69.02), followed by TACLE (71.43 and 72.81). On ImageNet-R, LDC leads by 5.4 and 7.1 points (38.08 and 45.78 vs. 32.66 and 38.67), followed by SLCA (37.71 and 45.50). MAGIC's accuracy grows little with the label ratio: on CIFAR-100 from 63.59 at 1% to 69.02 at 10%, against 63.36 to 75.29 for SLCA. Its backbone is frozen, so extra labels cannot improve its features, and its gain from unlabeled data shrinks as labels grow (on CIFAR-100 from 13.2 points at 1% to 2.5 at 10%; Section 5.3). On ImageNet-R the frozen ImageNet features also make the graph's labels unreliable: at 1%, only 21% of the labels that the graph gives the unlabeled images are correct, against 79% on CIFAR-100 (Section 5.4).

MAGIC achieves this without storing exemplars: when labels are scarce it outperforms the replay-based USP (at 1%: 63.59, 37.32 and 19.78 against 47.55, 22.83 and 17.50). It also uses a single configuration for all datasets and label ratios, whereas the baselines run with their published per-dataset settings.

Table 6 confirms these patterns setting by setting. In AIA, MAGIC is ahead of all eleven baselines on CUB-200 at 1% and 5% and of nine at 10%; at 1% on ImageNet-R it is ahead of nine and cannot be separated from the other two; at 1% on CIFAR-100 it is ahead of eight, level with two and behind TACLE; at 5% and 10% on CIFAR-100 and ImageNet-R, between two and five baselines are ahead of it. In last-task accuracy, no baseline is ahead of MAGIC at 1% on any dataset or at any ratio on CUB-200, although TACLE's mean on CIFAR-100 at 1% is 1.8 points higher. Because each baseline was run only three times, on labeled subsets other than MAGIC's, these comparisons are unpaired and their intervals are wide: the interval of the 8.9-point lead on CUB-200 at 1% runs from +2.7 to +15.2 points.

All methods were run with one class order, the native one. For MAGIC we also measured how much the results depend on the label split and on the class order (Appendix C). Over five label splits under each of three class orders, its AIA varies with a standard deviation of 0.7 to 1.9 points, and another class order alone shifts its mean AIA by up to 3.0 points. The baselines were not run under other class orders (Section 5.6), so a difference of this size between two methods, such as that between MAGIC and SLCA at 1% on ImageNet-R, should not be read as a ranking.

**MAGIC's forgetting is low where labels are scarce** (Table 5). At 1% it is the lowest of all methods on CIFAR-100 (9.8 points) and the second lowest on ImageNet-R (2.8). On CUB-200 and ImageNet-R it stays between 2.8 and 7.1 points at every ratio, the lowest or second lowest value in five of these six settings. On CIFAR-100, however, its forgetting grows with the label ratio, from 9.8 points at 1% to 15.8 at 5% and 21.8 at 10%, where only LwF and PASS forget more. More labels raise the accuracy that a task reaches when it is learned (first-task accuracy 78.1% at 1% and 89.1% at 10%) much more than the accuracy that survives to the end (54.4% and 57.4%), and forgetting measures the difference. Forgetting must also be read together with accuracy, since a method that learns little forgets little; at 1%, MAGIC combines low forgetting with the highest last-task accuracy on CUB-200 and ImageNet-R and the second highest on CIFAR-100. Its stability comes from its design: the backbone is frozen, the class statistics are stored in the frozen feature space, where they cannot drift, and the phantom features replayed from them keep the old classes in the classifier. Without replay, its forgetting on CIFAR-100 at 1% rises from 9.8 to 68.6 points (Section 5.5). Backbone-training baselines such as LwF and PASS forget far more (up to 91.0 and 33.8 points on CIFAR-100).

TACLE's results depend strongly on the number of labeled images per class (Table 1). On CIFAR-100, where every class has at least five, it is the strongest method at 1% (65.76) and second to SLCA at 5% and 10% (71.43 and 72.81), ahead of MAGIC by 2.2, 3.4 and 3.8 points: its confidence gate works there (Section 5.3), and it adapts the backbone, which MAGIC does not. Its accuracy nevertheless falls well below the numbers reported by its authors, and with one labeled image per class it fails. Their published results use a pre-trained ViT backbone, whose features are substantially more separable than the ResNet-18 representation we adopt to ensure cross-method comparability (Section 5.1.4). TACLE's selection pipeline (a task-adaptive confidence threshold combined with class-aware pseudo-labeling) relies on that separability: under a weaker representation, the hard accept/reject decision admits more incorrect pseudo-labels. Consequently, at $r = 1\%$ on fine-grained CUB-200 TACLE collapses to 9.94, below even the frozen-backbone baselines trained on the labeled images alone and 11.2 points below its own labeled-only variant (Section 5.3). Section 5.4 shows the same failure inside MAGIC: with pseudo-labels from the model's own predictions and a fixed threshold, AIA on CUB-200 at 1% falls 4.7 points below training on the labeled images alone.

Figure 2 plots the accuracy on all seen classes after each task, $A_t$, at $r = 1\%$ for six representative methods. **MAGIC's advantage comes from a flatter curve, not from a higher start.** On CIFAR-100 and ImageNet-R it starts below SLCA (78.1% and 25.4% after the first task, against 83.8% and 28.1%), but it loses accuracy more slowly: from the first task to the last it drops 23.7 points on CIFAR-100 and 8.5 on ImageNet-R, against 33.0 and 12.7 for SLCA. It overtakes SLCA at the fourth task on CIFAR-100 and at the sixth on ImageNet-R and finishes above it (54.4% vs. 50.8%, and 16.9% vs. 15.3%), with the highest final accuracy of all methods on ImageNet-R. On CIFAR-100, TACLE starts highest (86.6%) and also finishes highest (56.2%); MAGIC draws level with it from the fifth to the seventh task (within 0.4 points) and ends 1.8 points below it. On CUB-200, MAGIC leads from the first task (50.6% vs. 45.1% for SLCA) and ends at 29.2%, against 15.0% for the next methods (SLCA and USP), while TACLE and LDC end at 7.0% and 6.1%.

These curves show the stability–plasticity trade-off of Table 3 at the level of single tasks. Because the backbone is frozen and the class statistics are stored in its feature space, MAGIC keeps its old classes as new ones arrive, so a lower first-task fit ends as the best final accuracy on ImageNet-R and the second best on CIFAR-100. TACLE's curves show the two sides of its confidence gate: on CIFAR-100 it has the highest accuracy after the first task and after the last, whereas on CUB-200 it reaches only 12.4% after the first task, against 26.5–50.6% for the other five methods, and stays lowest or second lowest to the end.

Appendix B (Figure B1) shows the same curves at $r = 5\%$ and 10%. There, on CIFAR-100 and ImageNet-R, the strongest backbone-adapting baselines start higher and stay ahead to the last task, the per-task form of the crossover described above.

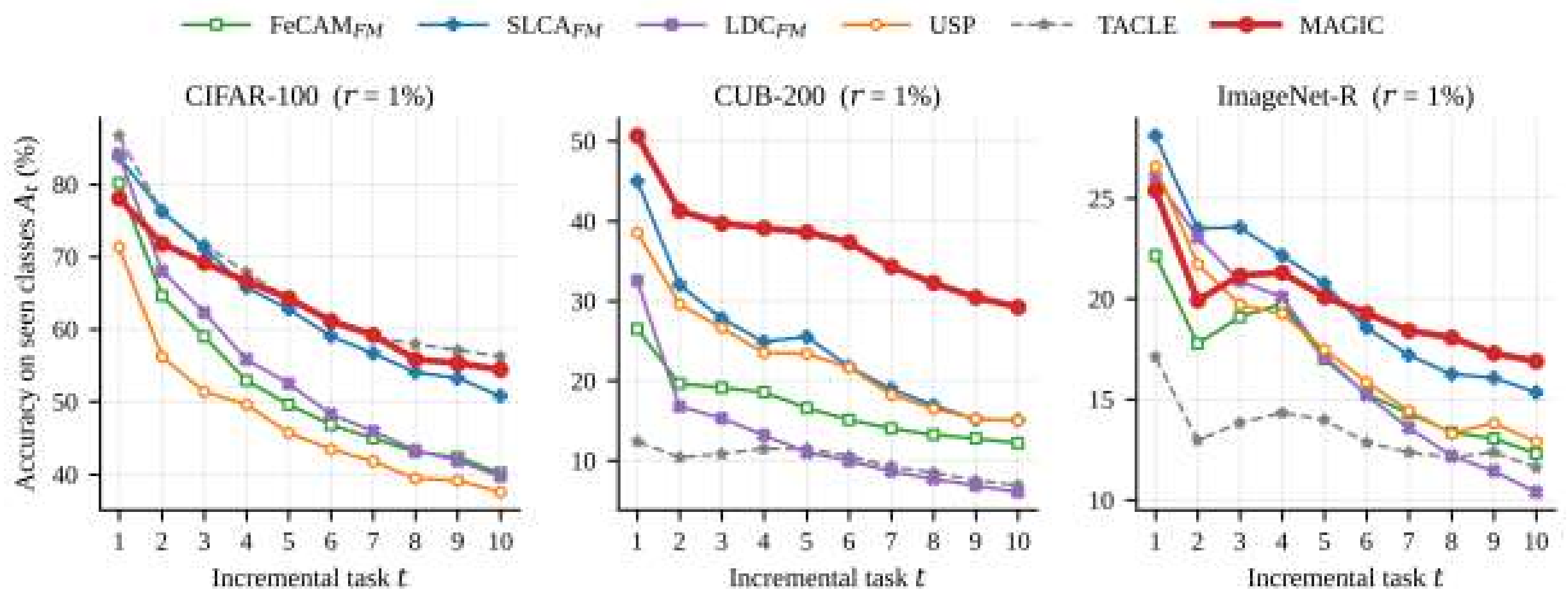

**Figure 2.** *Per-task accuracy $A_t$ (accuracy on all seen classes) for $r = 1\%$ across CIFAR-100, CUB-200, and ImageNet-R, averaged over seeds (three for the baselines, five for MAGIC) for six representative methods.*

**Table 3.** *Average incremental accuracy (AIA, %↑) in the semi-supervised setting, across three datasets and three label ratios. Mean ± std. over seeds, three for the baselines and five for MAGIC (std. shown as subscript);* ***Bold****: best per column; underline: second best. †: replay-based (all others are exemplar-free). The 100% column is each supervised method's fully-supervised upper bound (its original all-labels setting); it is not defined for the semi-supervised methods (USP, TACLE, MAGIC), marked –.*

| **Method** | **CIFAR-100** | | | | **CUB-200** | | | | **ImageNet-R** | | | |
|---|---|---|---|---|---|---|---|---|---|---|---|---|
| | **1%** | **5%** | **10%** | **100%** | **1%** | **5%** | **10%** | **100%** | **1%** | **5%** | **10%** | **100%** |
| *Frozen-backbone baselines (+ FixMatch at 1/5/10%; native fully-supervised at 100%)* | | | | | | | | | | | | |
| **SimpleCIL$_{FM}$** | 49.92$_{\pm 1.11}$ | 61.37$_{\pm 0.49}$ | 65.00$_{\pm 0.49}$ | 67.57$_{\pm 0.08}$ | 26.13$_{\pm 0.67}$ | 33.81$_{\pm 0.50}$ | 37.12$_{\pm 2.74}$ | 63.28$_{\pm 0.27}$ | 14.57$_{\pm 0.82}$ | 27.83$_{\pm 1.11}$ | 32.60$_{\pm 0.67}$ | 43.48$_{\pm 0.66}$ |
| **APER$_{FM}$** | 55.65$_{\pm 1.85}$ | 66.05$_{\pm 0.25}$ | 68.39$_{\pm 0.55}$ | 71.83$_{\pm 0.22}$ | 28.41$_{\pm 2.84}$ | 40.28$_{\pm 1.67}$ | 45.77$_{\pm 1.82}$ | 65.73$_{\pm 0.28}$ | 18.30$_{\pm 0.08}$ | 32.49$_{\pm 1.03}$ | 38.95$_{\pm 0.67}$ | 51.69$_{\pm 0.39}$ |
| **FeTrIL$_{FM}$** | 59.69$_{\pm 2.07}$ | 67.38$_{\pm 1.37}$ | 69.69$_{\pm 0.26}$ | 73.24$_{\pm 0.04}$ | 22.82$_{\pm 2.20}$ | 31.99$_{\pm 1.74}$ | 40.83$_{\pm 0.42}$ | 65.81$_{\pm 0.18}$ | 17.74$_{\pm 0.85}$ | 32.56$_{\pm 1.49}$ | 37.80$_{\pm 0.70}$ | 48.80$_{\pm 0.15}$ |
| **FeCAM$_{FM}$** | 52.34$_{\pm 0.11}$ | 67.86$_{\pm 0.32}$ | 70.71$_{\pm 2.88}$ | **80.65$_{\pm 0.49}$** | 16.76$_{\pm 0.98}$ | 26.70$_{\pm 5.97}$ | 40.47$_{\pm 1.78}$ | **67.53$_{\pm 0.54}$** | 16.40$_{\pm 0.34}$ | 35.18$_{\pm 2.01}$ | 44.92$_{\pm 1.18}$ | 59.62$_{\pm 0.16}$ |
| **RanPAC$_{FM}$** | 52.16$_{\pm 1.10}$ | 67.23$_{\pm 1.48}$ | 71.24$_{\pm 1.63}$ | 80.37$_{\pm 0.04}$ | 20.31$_{\pm 2.33}$ | 37.15$_{\pm 1.73}$ | 42.30$_{\pm 0.95}$ | 67.45$_{\pm 0.31}$ | 15.37$_{\pm 0.89}$ | 33.27$_{\pm 0.61}$ | 39.66$_{\pm 0.47}$ | 54.24$_{\pm 0.56}$ |
| *Backbone-training baselines (+ FixMatch at 1/5/10%; native fully-supervised at 100%)* | | | | | | | | | | | | |
| **LwF$_{FM}$** | 23.27$_{\pm 0.46}$ | 26.56$_{\pm 0.39}$ | 27.02$_{\pm 0.40}$ | 27.33$_{\pm 0.37}$ | 2.11$_{\pm 0.89}$ | 8.36$_{\pm 2.02}$ | 10.58$_{\pm 0.82}$ | 23.06$_{\pm 0.07}$ | 6.53$_{\pm 2.35}$ | 14.39$_{\pm 0.62}$ | 18.02$_{\pm 2.04}$ | 20.16$_{\pm 0.06}$ |
| **PASS$_{FM}$** | 33.63$_{\pm 3.12}$ | 51.06$_{\pm 1.30}$ | 56.81$_{\pm 1.34}$ | 65.26$_{\pm 0.25}$ | 7.59$_{\pm 2.12}$ | 17.73$_{\pm 0.66}$ | 21.41$_{\pm 1.28}$ | 45.82$_{\pm 0.94}$ | 7.99$_{\pm 0.58}$ | 22.72$_{\pm 0.61}$ | 31.26$_{\pm 0.97}$ | 46.66$_{\pm 1.34}$ |
| **SLCA$_{FM}$** | 63.36$_{\pm 1.59}$ | **73.14$_{\pm 0.19}$** | **75.29$_{\pm 0.04}$** | 77.79$_{\pm 0.23}$ | 24.31$_{\pm 1.97}$ | 35.95$_{\pm 1.06}$ | 41.31$_{\pm 0.61}$ | 63.03$_{\pm 0.18}$ | **20.14$_{\pm 0.36}$** | 37.71$_{\pm 1.13}$ | 45.50$_{\pm 0.63}$ | 56.19$_{\pm 0.22}$ |
| **LDC$_{FM}$** | 54.17$_{\pm 0.84}$ | 66.96$_{\pm 1.99}$ | 69.67$_{\pm 1.47}$ | 76.89$_{\pm 0.07}$ | 12.82$_{\pm 0.60}$ | 27.06$_{\pm 2.53}$ | 40.64$_{\pm 1.03}$ | 60.46$_{\pm 0.05}$ | 17.00$_{\pm 0.99}$ | **38.08$_{\pm 0.41}$** | **45.78$_{\pm 1.32}$** | **60.21$_{\pm 0.19}$** |
| *Native SSCIL (semi-supervised only — no fully-supervised mode)* | | | | | | | | | | | | |
| **USP†** | 47.55$_{\pm 2.04}$ | 61.49$_{\pm 1.11}$ | 63.54$_{\pm 1.26}$ | – | 22.83$_{\pm 0.81}$ | 34.86$_{\pm 1.18}$ | 42.81$_{\pm 0.48}$ | – | 17.50$_{\pm 1.45}$ | 35.50$_{\pm 1.91}$ | 42.56$_{\pm 0.99}$ | – |
| **TACLE** | **65.76$_{\pm 0.82}$** | 71.43$_{\pm 1.17}$ | 72.81$_{\pm 0.33}$ | – | 9.94$_{\pm 1.75}$ | 32.68$_{\pm 4.24}$ | 42.69$_{\pm 5.06}$ | – | 13.34$_{\pm 0.44}$ | 31.22$_{\pm 0.85}$ | 38.48$_{\pm 0.22}$ | – |
| **MAGIC** | 63.59$_{\pm 0.97}$ | 68.05$_{\pm 0.20}$ | 69.02$_{\pm 0.12}$ | – | **37.32$_{\pm 1.22}$** | **45.19$_{\pm 0.97}$** | **47.77$_{\pm 0.92}$** | – | 19.78$_{\pm 0.46}$ | 32.66$_{\pm 0.89}$ | 38.67$_{\pm 0.51}$ | – |

**Table 4.** *Last task accuracy ($A_T$, %↑) across three datasets and three label ratios. Mean ± std. over seeds, three for the baselines and five for MAGIC (std. shown as subscript); **Bold**: best per column; underline: second best. †: replay-based (all others are exemplar-free). The 100% column is each supervised method's fully-supervised upper bound (its original all-labels setting); it is not defined for the semi-supervised methods (USP, TACLE, MAGIC), marked –.*

| Method | CIFAR-100 | | | | CUB-200 | | | | ImageNet-R | | | |
|---|---|---|---|---|---|---|---|---|---|---|---|---|
| | 1% | 5% | 10% | 100% | 1% | 5% | 10% | 100% | 1% | 5% | 10% | 100% |
| *Frozen-backbone baselines (+ FixMatch at 1/5/10%; native fully-supervised at 100%)* | | | | | | | | | | | | |
| **SimpleCIL$_{FM}$** | 36.51$_{\pm1.03}$ | 47.24$_{\pm0.52}$ | 50.96$_{\pm0.10}$ | 55.78$_{\pm0.06}$ | 15.99$_{\pm0.78}$ | 20.55$_{\pm0.59}$ | 23.17$_{\pm1.87}$ | 54.37$_{\pm0.27}$ | 9.62$_{\pm0.98}$ | 20.54$_{\pm1.03}$ | 25.25$_{\pm0.28}$ | 36.76$_{\pm0.51}$ |
| **APER$_{FM}$** | 42.45$_{\pm0.90}$ | 54.02$_{\pm0.16}$ | 56.24$_{\pm0.72}$ | 59.77$_{\pm0.22}$ | 19.29$_{\pm1.64}$ | 28.88$_{\pm0.94}$ | 34.52$_{\pm0.69}$ | 57.45$_{\pm0.10}$ | 13.62$_{\pm0.40}$ | 25.86$_{\pm0.17}$ | 32.26$_{\pm0.34}$ | 43.26$_{\pm0.35}$ |
| **FeTrIL$_{FM}$** | 46.97$_{\pm0.59}$ | 54.72$_{\pm0.03}$ | 57.31$_{\pm0.20}$ | 63.04$_{\pm0.08}$ | 15.17$_{\pm1.95}$ | 25.20$_{\pm1.67}$ | 31.56$_{\pm0.14}$ | 58.13$_{\pm0.16}$ | 13.84$_{\pm0.42}$ | 26.77$_{\pm1.10}$ | 30.93$_{\pm0.67}$ | 42.95$_{\pm0.10}$ |
| **FeCAM$_{FM}$** | 40.19$_{\pm0.27}$ | 55.53$_{\pm0.38}$ | 59.19$_{\pm3.50}$ | **72.04$_{\pm0.76}$** | 12.17$_{\pm0.59}$ | 18.96$_{\pm5.38}$ | 30.24$_{\pm1.61}$ | 59.01$_{\pm0.57}$ | 12.29$_{\pm0.26}$ | 28.84$_{\pm1.42}$ | 37.00$_{\pm0.54}$ | 50.38$_{\pm0.21}$ |
| **RanPAC$_{FM}$** | 39.53$_{\pm0.80}$ | 56.25$_{\pm1.00}$ | 60.23$_{\pm1.46}$ | 71.32$_{\pm0.23}$ | 15.48$_{\pm0.76}$ | 27.19$_{\pm1.40}$ | 32.49$_{\pm0.89}$ | **60.28$_{\pm0.09}$** | 14.83$_{\pm0.87}$ | 27.48$_{\pm0.67}$ | 33.12$_{\pm0.35}$ | 47.52$_{\pm0.30}$ |
| *Backbone-training baselines (+ FixMatch at 1/5/10%; native fully-supervised at 100%)* | | | | | | | | | | | | |
| **LwF$_{FM}$** | 9.20$_{\pm0.22}$ | 9.67$_{\pm0.44}$ | 9.59$_{\pm0.07}$ | 9.63$_{\pm0.13}$ | 0.28$_{\pm0.46}$ | 2.66$_{\pm0.35}$ | 3.59$_{\pm0.53}$ | 8.03$_{\pm0.07}$ | 1.67$_{\pm0.83}$ | 5.72$_{\pm0.64}$ | 7.23$_{\pm0.98}$ | 8.70$_{\pm0.10}$ |
| **PASS$_{FM}$** | 25.26$_{\pm2.78}$ | 34.26$_{\pm0.25}$ | 40.64$_{\pm5.17}$ | 45.41$_{\pm2.62}$ | 5.10$_{\pm1.43}$ | 10.98$_{\pm0.24}$ | 13.88$_{\pm1.17}$ | 19.18$_{\pm1.45}$ | 4.75$_{\pm0.77}$ | 16.48$_{\pm0.82}$ | 23.91$_{\pm1.00}$ | 27.70$_{\pm0.71}$ |
| **SLCA$_{FM}$** | 50.76$_{\pm0.96}$ | **61.38$_{\pm0.11}$** | **64.04$_{\pm0.16}$** | 66.85$_{\pm0.12}$ | 15.04$_{\pm0.58}$ | 24.79$_{\pm1.45}$ | 30.21$_{\pm0.92}$ | 54.39$_{\pm0.49}$ | 15.34$_{\pm0.70}$ | **31.41$_{\pm0.59}$** | 37.94$_{\pm0.04}$ | 48.15$_{\pm0.34}$ |
| **LDC$_{FM}$** | 39.74$_{\pm0.71}$ | 54.33$_{\pm1.26}$ | 57.79$_{\pm0.28}$ | 65.19$_{\pm0.22}$ | 6.10$_{\pm0.41}$ | 15.09$_{\pm1.85}$ | 26.42$_{\pm0.86}$ | 50.83$_{\pm0.31}$ | 10.40$_{\pm0.19}$ | 26.95$_{\pm0.32}$ | 34.62$_{\pm1.55}$ | **51.24$_{\pm0.57}$** |
| *Native SSCIL (semi-supervised only — no fully-supervised mode)* | | | | | | | | | | | | |
| **USP†** | 37.56$_{\pm2.19}$ | 51.50$_{\pm0.99}$ | 52.77$_{\pm0.29}$ | – | 15.00$_{\pm0.98}$ | 26.49$_{\pm2.11}$ | 34.29$_{\pm0.48}$ | – | 12.88$_{\pm1.44}$ | 30.34$_{\pm5.54}$ | **38.06$_{\pm0.71}$** | – |
| **TACLE** | **56.20$_{\pm0.94}$** | 60.78$_{\pm0.34}$ | 61.76$_{\pm0.96}$ | – | 6.95$_{\pm2.23}$ | 21.98$_{\pm4.95}$ | 31.22$_{\pm5.87}$ | – | 11.62$_{\pm1.24}$ | 25.57$_{\pm1.02}$ | 31.35$_{\pm0.46}$ | – |
| **MAGIC** | 54.38$_{\pm0.48}$ | 57.15$_{\pm0.45}$ | 57.36$_{\pm0.16}$ | – | **29.25$_{\pm0.65}$** | **35.89$_{\pm0.46}$** | **39.00$_{\pm0.82}$** | – | **16.88$_{\pm0.51}$** | 28.92$_{\pm0.46}$ | 34.40$_{\pm0.45}$ | – |

**Table 5.** *Average forgetting ($F_T$, %↓) across three datasets and three label ratios. Mean ± std. over seeds, three for the baselines and five for MAGIC (std. shown as subscript);* ***Bold****: lowest per column; underline: second lowest. †: replay-based (all others are exemplar-free). The 100% column is each supervised method's fully-supervised upper bound (its original all-labels setting); it is not defined for the semi-supervised methods (USP, TACLE, MAGIC), marked –.*

| Method | CIFAR-100 | | | | CUB-200 | | | | ImageNet-R | | | |
|---|---|---|---|---|---|---|---|---|---|---|---|---|
| | 1% | 5% | 10% | 100% | 1% | 5% | 10% | 100% | 1% | 5% | 10% | 100% |
| *Frozen-backbone baselines (+ FixMatch at 1/5/10%; native fully-supervised at 100%)* | | | | | | | | | | | | |
| **SimpleCIL$_{FM}$** | 14.78$_{\pm0.49}$ | 15.88$_{\pm0.87}$ | 14.42$_{\pm1.51}$ | 12.62$_{\pm0.06}$ | 7.21$_{\pm0.92}$ | 11.45$_{\pm0.59}$ | 11.38$_{\pm0.60}$ | 8.46$_{\pm0.02}$ | 5.61$_{\pm0.37}$ | 7.97$_{\pm0.73}$ | 9.19$_{\pm0.61}$ | 8.21$_{\pm0.59}$ |
| **APER$_{FM}$** | 13.94$_{\pm0.83}$ | 13.41$_{\pm0.38}$ | 12.89$_{\pm0.79}$ | 11.67$_{\pm0.79}$ | 8.48$_{\pm0.42}$ | 8.50$_{\pm0.68}$ | 9.22$_{\pm0.95}$ | 7.88$_{\pm0.18}$ | 4.78$_{\pm0.20}$ | 6.85$_{\pm0.35}$ | 7.39$_{\pm0.23}$ | 8.54$_{\pm0.20}$ |
| **FeTrIL$_{FM}$** | 20.82$_{\pm2.35}$ | 20.23$_{\pm1.42}$ | 20.55$_{\pm0.65}$ | 21.27$_{\pm0.13}$ | 7.46$_{\pm1.16}$ | **6.39$_{\pm0.59}$** | 7.68$_{\pm0.11}$ | 10.49$_{\pm0.09}$ | 3.07$_{\pm0.94}$ | 5.87$_{\pm1.00}$ | 8.94$_{\pm0.13}$ | 11.06$_{\pm0.04}$ |
| **FeCAM$_{FM}$** | 13.99$_{\pm0.85}$ | 12.99$_{\pm0.43}$ | 11.86$_{\pm1.45}$ | **8.89$_{\pm0.20}$** | 7.00$_{\pm2.86}$ | 10.31$_{\pm1.58}$ | 11.46$_{\pm0.54}$ | 7.86$_{\pm0.18}$ | 4.87$_{\pm0.14}$ | 7.96$_{\pm0.02}$ | 8.25$_{\pm0.30}$ | 6.87$_{\pm0.11}$ |
| **RanPAC$_{FM}$** | 13.92$_{\pm1.03}$ | 12.29$_{\pm0.54}$ | 12.04$_{\pm0.43}$ | 9.57$_{\pm0.27}$ | 3.94$_{\pm2.04}$ | 9.55$_{\pm0.52}$ | 9.63$_{\pm0.58}$ | 7.06$_{\pm0.11}$ | 3.35$_{\pm1.27}$ | 6.10$_{\pm0.87}$ | 7.10$_{\pm0.21}$ | 7.71$_{\pm0.83}$ |
| *Backbone-training baselines (+ FixMatch at 1/5/10%; native fully-supervised at 100%)* | | | | | | | | | | | | |
| **LwF$_{FM}$** | 78.43$_{\pm1.68}$ | 89.06$_{\pm0.64}$ | 91.03$_{\pm1.16}$ | 93.19$_{\pm1.35}$ | 10.82$_{\pm4.12}$ | 23.90$_{\pm3.37}$ | 31.95$_{\pm0.10}$ | 74.07$_{\pm0.65}$ | 16.66$_{\pm6.65}$ | 40.05$_{\pm2.57}$ | 49.95$_{\pm0.21}$ | 67.52$_{\pm0.20}$ |
| **PASS$_{FM}$** | 28.16$_{\pm2.94}$ | 33.62$_{\pm1.10}$ | 33.82$_{\pm1.14}$ | 32.97$_{\pm3.33}$ | 11.52$_{\pm0.54}$ | 21.34$_{\pm2.00}$ | 23.37$_{\pm0.67}$ | 48.13$_{\pm0.99}$ | 10.90$_{\pm0.76}$ | 16.04$_{\pm0.71}$ | 15.77$_{\pm0.80}$ | 22.50$_{\pm0.36}$ |
| **SLCA$_{FM}$** | 20.28$_{\pm0.13}$ | 13.14$_{\pm0.17}$ | 11.45$_{\pm0.41}$ | 11.50$_{\pm0.08}$ | 17.81$_{\pm0.79}$ | 15.04$_{\pm0.21}$ | 13.56$_{\pm1.06}$ | **6.57$_{\pm0.56}$** | 7.50$_{\pm0.34}$ | 7.32$_{\pm0.62}$ | 6.96$_{\pm0.38}$ | **4.24$_{\pm0.08}$** |
| **LDC$_{FM}$** | 12.20$_{\pm0.14}$ | 15.38$_{\pm1.65}$ | 16.44$_{\pm1.51}$ | 16.21$_{\pm0.29}$ | 4.19$_{\pm0.46}$ | 9.20$_{\pm0.65}$ | 11.76$_{\pm0.47}$ | 17.51$_{\pm0.39}$ | 3.80$_{\pm0.53}$ | 8.67$_{\pm0.94}$ | 10.41$_{\pm0.65}$ | 12.27$_{\pm0.12}$ |
| *Native SSCIL (semi-supervised only — no fully-supervised mode)* | | | | | | | | | | | | |
| **USP†** | 10.95$_{\pm1.47}$ | **10.18$_{\pm0.75}$** | **11.11$_{\pm0.98}$** | – | 7.95$_{\pm0.55}$ | 8.37$_{\pm0.50}$ | 8.82$_{\pm0.57}$ | – | 4.55$_{\pm0.21}$ | **3.34$_{\pm0.25}$** | **3.28$_{\pm0.78}$** | – |
| **TACLE** | 12.05$_{\pm0.55}$ | 12.37$_{\pm0.46}$ | 12.25$_{\pm1.47}$ | – | **3.29$_{\pm3.00}$** | 7.44$_{\pm4.39}$ | 10.16$_{\pm5.57}$ | – | **1.53$_{\pm0.33}$** | 5.58$_{\pm0.36}$ | 7.18$_{\pm0.27}$ | – |
| **MAGIC** | **9.81$_{\pm0.38}$** | 15.75$_{\pm0.87}$ | 21.76$_{\pm0.91}$ | – | 5.72$_{\pm0.42}$ | 6.48$_{\pm0.67}$ | **7.13$_{\pm0.35}$** | – | 2.83$_{\pm0.22}$ | 4.57$_{\pm0.33}$ | 4.68$_{\pm0.62}$ | – |

**Table 6.** *MAGIC against the baselines in each setting. Strongest: the baseline with the highest mean. Δ: MAGIC minus the strongest baseline, with the 95% confidence interval of Welch's t-test (five MAGIC runs, three baseline runs). Ahead / level / behind: the number of the eleven baselines whose interval for MAGIC minus baseline lies above zero, contains zero, or lies below zero.*

| | AIA | | | Last-task accuracy $A_T$ | | |
|---|---|---|---|---|---|---|
| **Labels** | **Strongest** | **Δ [95% CI]** | **Ahead / level / behind** | **Strongest** | **Δ [95% CI]** | **Ahead / level / behind** |
| *CIFAR-100* | | | | | | |
| 1% | TACLE (65.76) | −2.17 [−3.82, −0.52] | 8 / 2 / 1 | TACLE (56.20) | −1.82 [−3.82, +0.19] | 10 / 1 / 0 |
| 5% | SLCA (73.14) | −5.09 [−5.47, −4.72] | 5 / 4 / 2 | SLCA (61.38) | −4.23 [−4.78, −3.68] | 7 / 2 / 2 |
| 10% | SLCA (75.29) | −6.27 [−6.41, −6.13] | 4 / 4 / 3 | SLCA (64.04) | −6.68 [−6.99, −6.36] | 4 / 5 / 2 |
| *CUB-200* | | | | | | |
| 1% | APER (28.41) | +8.91 [+2.65, +15.17] | 11 / 0 / 0 | APER (19.29) | +9.96 [+6.29, +13.63] | 11 / 0 / 0 |
| 5% | APER (40.28) | +4.91 [+1.43, +8.39] | 11 / 0 / 0 | APER (28.88) | +7.01 [+4.98, +9.03] | 11 / 0 / 0 |
| 10% | APER (45.77) | +2.00 [−1.90, +5.89] | 9 / 2 / 0 | APER (34.52) | +4.48 [+3.09, +5.87] | 10 / 1 / 0 |
| *ImageNet-R* | | | | | | |
| 1% | SLCA (20.14) | −0.36 [−1.10, +0.38] | 9 / 2 / 0 | SLCA (15.34) | +1.54 [+0.14, +2.95] | 11 / 0 / 0 |
| 5% | LDC (38.08) | −5.42 [−6.56, −4.28] | 3 / 6 / 2 | SLCA (31.41) | −2.49 [−3.66, −1.31] | 7 / 3 / 1 |
| 10% | LDC (45.78) | −7.11 [−10.07, −4.15] | 3 / 3 / 5 | USP (38.06) | −3.66 [−5.12, −2.20] | 7 / 1 / 3 |

### 5.3. When does unlabeled data help?

In this subsection, we measure how much accuracy each method recovers from the unlabeled stream. This tests the premise of Section 1: the value of unlabeled data in incremental learning depends on how it is selected and weighted, not only on its presence. For each method and setting, Table 7 reports $\Delta$ = AIA(semi-supervised) − AIA(labeled-only), where the labeled-only counterpart discards the unlabeled stream and keeps everything else fixed.

The labeled-only variants are obtained differently for different methods. For the retrofitted baselines, we switch off the FixMatch objective, which leaves the underlying supervised learner untouched. For the native semi-supervised approaches, we rely on each method's own supervised reduction. TACLE's training pipeline already includes a fully-supervised branch, invoked automatically when no unlabeled loader is supplied, which estimates Gaussian class statistics from labeled features alone; we adopt that branch without modification. For USP, we zero out the weights of all unlabeled objectives (the divide-and-conquer pseudo-label loss, plus the unlabeled feature-structure, distillation and drift terms), which keeps only its supervised cross-entropy, labeled feature-structure regularization and exemplar-based classification. MAGIC's labeled-only variant uses no unlabeled image at all: it drops the unlabeled loss and computes its class statistics from the labeled images only.

The dominant finding is that the sign and magnitude of $\Delta$ depend far more on the host method than on the volume of unlabeled data. Across the eleven baselines, $\Delta$ ranges from −11.2 to +36.2, and 26 out of 99 entries are negative. SimpleCIL loses accuracy in every setting (−0.4 to −7.4).

Since its class prototypes are plain feature means, any confidence-gated pseudo-label that lands on the wrong prototype contaminates the class mean; this effect is most severe on the fine-grained CUB-200. LDC records the largest raw gains (+5.4 to +36.2), but this number is deceptive: its labeled-only model collapses outright, because backbone training on the $r\%$ subset fails, and FixMatch acts as life support rather than refinement. Despite its large $\Delta$, LDC's absolute accuracy at $r = 1\%$ remains below MAGIC's, so the gain must be read together with the absolute score. The other baselines fall in between, and their behavior tracks how each method uses statistics. FeTrIL's $\Delta$ declines steadily as $r$ increases (from +6.4 to +0.5 on CIFAR-100) because its labeled statistics saturate quickly. FeCAM follows the opposite trend: its $\Delta$ grows with $r$ on ImageNet-R (+2.9 to +7.4), since covariance estimation benefits from additional pseudo-labeled samples, and those samples become more reliable as the labeled subset grows.

The two natively semi-supervised methods behave quite differently. USP gains almost nothing from the unlabeled stream (−0.3 to +3.4, within seed variability in five of nine settings); its strong standing in Tables 3–5 comes from its exemplar replay and supervised components, not from unlabeled data. TACLE's threshold pipeline extracts substantial value when confidence gating works: on CIFAR-100, where every class has five labeled images even at $r = 1\%$, its gains (+26.0 → +8.2 → +6.0) exceed MAGIC's at every ratio (+13.2 → +4.1 → +2.5), and so does its AIA (Table 3). But it fails in precisely the settings where semi-supervised learning should help most: on CUB-200 and ImageNet-R its gains appear only at 5% and 10% labels (−11.2 → +3.3 → +9.6 and +0.1 → +5.4 → +5.2), and at $r = 1\%$ on these fine-grained and distribution-shifted datasets the unlabeled data do harm or add nothing, exactly where MAGIC gains +14.8 and +5.7. With roughly one label per class, the pseudo-labels that pass the confidence gate do not help the threshold-based learner at the very moment it needs the unlabeled data most; on CUB-200 they damage it from the first task, after which its accuracy is 12.4%, against 48.9% for its labeled-only variant. The same failure appears in the hardest setting, CUB-200 at $r = 1\%$, where six of the eleven baselines lose accuracy and the median gain is negative (−1.4).

MAGIC, whose row reflects its native propagation-and-SWGC pathway rather than a FixMatch retrofit, gains in all nine settings, and the 95% interval of every gain, paired over the five seeds, lies above zero (Section 5.5). The gains are largest where supervision is thinnest (+13.2 on CIFAR-100 and +14.8 on CUB-200 at $r = 1\%$), fall to +2.5 and +10.0 at 10%, and range from +3.2 to +5.7 on ImageNet-R. A consistent sign alone is not unique: SLCA and LDC also gain in all nine settings. What sets MAGIC apart is the combination of three properties: gains in every setting, gains that are largest where labels are scarcest, and strong absolute accuracy; every other method has at most two of them. SLCA gains everywhere, but at $r = 1\%$ less than MAGIC on every dataset (+9.3, +1.1 and +1.9 against +13.2, +14.8 and +5.7). LDC's large gains mostly compensate for a collapsed labeled-only model, TACLE's gains on CUB-200 and ImageNet-R appear only at higher label ratios, and USP keeps its accuracy through exemplar replay rather than through unlabeled data. Entries with $|\Delta|$ below about 1 lie within seed variability and are inconclusive.

Taken together with Section 5.2, these results support the paper's practical takeaway: confidence-threshold SSL attached to an incremental learner is unreliable; it can be harmful, inert or indispensable depending on the host, and even a native threshold-based SSCIL method, whose gains on CIFAR-100 exceed MAGIC's, loses 11.2 points on CUB-200 and gains nothing on ImageNet-R at $r = 1\%$, where a class has fewer than two labeled images on average. Graph propagation, as used by MAGIC, gives consistent gains whose size grows with label scarcity, which is the case that semi-supervised incremental learning is meant to address.

**Table 7.** *Gain from unlabeled data, ΔAIA = (semi-supervised) − (labeled-only) (%↑). The labeled-only variant trains each method on the r% labeled subset with the unlabeled stream discarded; a positive Δ is the accuracy the method recovers from unlabeled data. MAGIC: mean over five seeds.*

| Method | CIFAR-100 | | | CUB-200 | | | ImageNet-R | | |
|---|---|---|---|---|---|---|---|---|---|
| | Δ@1% | Δ@5% | Δ@10% | Δ@1% | Δ@5% | Δ@10% | Δ@1% | Δ@5% | Δ@10% |
| *Frozen-backbone baselines* | | | | | | | | | |
| **SimpleCIL$_{\text{FM}}$** | −4.21 | −2.74 | −0.42 | −3.11 | −3.82 | −7.43 | −1.70 | −2.00 | −0.65 |
| **APER$_{\text{FM}}$** | 0.93 | 3.21 | 2.03 | −1.36 | −0.14 | −0.40 | 0.30 | 0.74 | 1.79 |
| **FeTrIL$_{\text{FM}}$** | 6.37 | 3.17 | 0.45 | 7.68 | 3.79 | 2.66 | 5.03 | 2.48 | −0.18 |
| **FeCAM$_{\text{FM}}$** | 6.44 | 6.45 | 4.41 | −5.47 | 0.22 | 1.40 | 2.85 | 4.03 | 7.35 |
| **RanPAC$_{\text{FM}}$** | −1.77 | −1.24 | −1.77 | 1.58 | 0.30 | 0.28 | 3.33 | 0.49 | −0.07 |
| *Backbone-training baselines* | | | | | | | | | |
| **LwF$_{\text{FM}}$** | 7.64 | 5.13 | 2.68 | −5.51 | −2.94 | −2.61 | 1.12 | 4.17 | 4.53 |
| **PASS$_{\text{FM}}$** | 1.36 | 1.77 | 1.60 | −4.18 | 2.76 | −0.30 | −2.14 | 1.73 | 4.11 |
| **SLCA$_{\text{FM}}$** | 9.29 | 7.44 | 3.14 | 1.11 | 1.73 | 2.33 | 1.91 | 5.18 | 4.92 |
| **LDC$_{\text{FM}}$** | 33.11 | 36.20 | 21.94 | 5.39 | 13.15 | 22.66 | 11.62 | 25.53 | 25.31 |
| *Native SSCIL* | | | | | | | | | |
| **USP** | 1.29 | 0.49 | 0.53 | 2.64 | 3.39 | 0.56 | 1.25 | −0.33 | 0.09 |
| **TACLE** | 26.01 | 8.17 | 5.99 | −11.21 | 3.34 | 9.58 | 0.11 | 5.42 | 5.21 |
| **MAGIC** | 13.19 | 4.11 | 2.47 | 14.84 | 13.47 | 10.03 | 5.71 | 4.21 | 3.24 |

## 5.4. Which pseudo-labels, and which gate? A comparison inside MAGIC

The baselines of Table 7 differ in much more than their pseudo-labels, so they cannot isolate what graph propagation contributes. We therefore keep MAGIC fixed and change only its pseudo-labeling rule, in both places where the rule acts: the unlabeled loss and the class statistics. A rule has two parts (Table 8). The *source* gives each unlabeled image a soft label: the graph (MAGIC), the model's own prediction (self-training, the source of FixMatch [45], FreeMatch [53] and SoftMatch [4]), or the most similar labeled image (nearest labeled; here labels do not travel through other unlabeled images). The *gate* decides how much each image counts: MAGIC's squared confidence, a fixed threshold $\tau = 0.95$ (FixMatch), FreeMatch's self-adaptive class-wise threshold, SoftMatch's truncated Gaussian weight, or weight 1 for every image. We take only the gates of FreeMatch and SoftMatch, without FreeMatch's fairness loss and SoftMatch's uniform alignment, and update their running statistics with momentum 0.99, because a task has only a few hundred training steps. MAGIC's gate keeps the whole soft label as the target; the other gates use the hard (arg-max) label, as in the original methods. As for the baselines (Section 5.1.3), the self-training and nearest-labeled rules choose among the classes of the current task, whereas graph labels range over all seen classes; the last paragraph of this section removes this difference. All rules run with the same five seeds as MAGIC, so every difference is paired.

**The source of the pseudo-labels matters most.** With every gate except the fixed threshold, graph labels give a higher AIA than self-training in all nine settings (Table 8). On CUB-200 at 1%, graph labels with MAGIC's weights reach 37.32, while self-training reaches 29.44 with the same weights and only 9.86–12.10 with the SoftMatch, FreeMatch or uniform gates, far below the 22.48 of the labeled images alone. Nearest-labeled pseudo-labels, which come from the labeled images only, reach at most 35.67 there: letting labels travel through the unlabeled images adds to what the labeled images alone say.

**A fixed high threshold removes most of the gain, whatever the source.** With graph labels and $\tau = 0.95$, AIA at 1% is 54.15, 22.72 and 14.22 on CIFAR-100, CUB-200 and ImageNet-R, close to the 50.40, 22.48 and 14.07 of the labeled images alone; on CUB-200 the threshold keeps 0.2 of the 14.8 points that unlabeled data add. The reason is coverage: with one to five labels per class, few images reach a confidence of 0.95, and during training the threshold keeps 3.9% of the unlabeled images on CIFAR-100, 0.9% on CUB-200 and 0.4% on ImageNet-R (Figure 3). With self-training, no fixed threshold from 0.3 to 0.95 even reaches the labeled-only model on CUB-200 and ImageNet-R (at best 20.66 and 11.88). Lower thresholds do work with graph labels: at $\tau = 0.3$ they reach 63.43, 37.58 and 21.11, level with or above MAGIC's weights (63.59, 37.32 and 19.78), but only with a threshold far below the usual values, found by sweeping it from 0.3 to 0.95 on these results.

**An adaptive threshold on the graph's labels works as well as MAGIC's weights, and better on CUB-200.** With FreeMatch's class-wise threshold, graph labels give the best AIA of all rules on CUB-200 at every ratio, 1.9, 2.2 and 2.0 points above MAGIC (intervals above zero), and at 1% on ImageNet-R (+1.5); on CIFAR-100 they stay within 0.8 points of MAGIC. FreeMatch keeps far more images than MAGIC's weights (coverage 0.72 against 0.17 on CUB-200 at 1%) at a lower accuracy (48% against 61%): on fine-grained data, more and noisier labels help the class statistics more than fewer and cleaner ones. So it is not the form of the weighting that makes MAGIC work, but the graph's labels, used through a gate that does not discard almost all of them. MAGIC keeps the squared confidence, which was fixed before these experiments; choosing the gate per dataset from these results would overstate its accuracy, and a self-adaptive gate on graph labels is a promising variant (Section 5.6).

Figure 3 shows why the source matters. For each rule at $r = 1\%$, it plots the accuracy of the pseudo-labels used in training against their coverage, the mean weight that an unlabeled image receives. **At the same coverage, graph labels are more accurate than the model's own.** Near a coverage of 0.6 on CIFAR-100, 89% of the graph's labels are correct against 85% of the model's; near 0.4 on CUB-200, 56% against 36%; near 0.2 on ImageNet-R, 48% against 25%. The model's confidence is a poor guide at these budgets: on CUB-200, raising the self-training threshold from 0.8 to 0.95 lowers the accuracy of the labels it keeps from 36% to 31%, whereas raising a threshold on the graph's confidence from 0.3 to 0.95 raises it from 56% to 89%.

These labels end up in the class statistics. **MAGIC's stored Gaussians are much closer to the true class distributions than those estimated from the labeled images alone.** We compare each stored mean and variance with those of all training images of the class under their true labels. At $r = 1\%$, the relative error of the stored means falls from 30% to 8% on CIFAR-100, from 55% to 21% on CUB-200 and from 64% to 29% on ImageNet-R, and the 2-Wasserstein distance between the phantom and the real features of a class falls from 8.7 to 1.9, from 19.1 to 5.9 and from 19.5 to 7.5. A fixed threshold of 0.95 on the graph's labels leaves the statistics on CUB-200 exactly as poor as the labeled images alone do (19.1), because almost no image passes it, and self-training with FixMatch's threshold improves them only slightly (17.0).

Finally, does a growing label space hurt confidence thresholds, as is often argued? In our protocol the baselines' pseudo-labels range over the classes of the current task only (Section 5.1.3), so their candidate set does not grow with the tasks. To test the mechanism directly, we let the self-training rules choose their pseudo-labels among all seen classes. At $r = 1\%$ this changes AIA by −4.6 to +2.1 points across the three gates we tested (fixed threshold, SoftMatch and squared confidence), and by $-1.6$, $-0.2$ and $-2.3$ points for FixMatch's threshold on CIFAR-100, CUB-200 and ImageNet-R. A growing label space therefore hurts a threshold somewhat, but it is not the main problem at these budgets; the main problem is that the model's own confidence says little about which of its labels are right.

**Table 8.** *Pseudo-labeling rules inside MAGIC: AIA (%) with MAGIC's rule replaced, mean over the same five seeds (std. at most 2.4). Source: graph (label propagation, as in MAGIC), self-training (the model's own prediction) or the nearest labeled image. Gate: squared confidence (MAGIC), fixed threshold $\tau = 0.95$ (FixMatch), FreeMatch's self-adaptive threshold, SoftMatch's truncated Gaussian weight, or weight 1. Each rule is used both in the unlabeled loss and in the class statistics. Bold: best per column; underline: second best.*

| | CIFAR-100 | | | CUB-200 | | | ImageNet-R | | |
|---|---|---|---|---|---|---|---|---|---|
| **Gate** | **1%** | **5%** | **10%** | **1%** | **5%** | **10%** | **1%** | **5%** | **10%** |
| *No unlabeled data* | | | | | | | | | |
| Labeled images only | 50.40 | 63.93 | 66.55 | 22.48 | 31.72 | 37.74 | 14.07 | 28.45 | 35.43 |
| *Graph labels (label propagation)* | | | | | | | | | |
| Squared confidence (MAGIC) | **63.59** | **68.05** | **69.02** | 37.32 | 45.19 | 47.77 | 19.78 | 32.66 | **38.67** |
| Fixed $\tau = 0.95$ | 54.15 | 63.04 | 65.53 | 22.72 | 31.66 | 37.71 | 14.22 | 28.41 | 35.17 |
| FreeMatch threshold | 63.41 | 67.78 | 68.23 | **39.17** | **47.36** | **49.72** | **21.31** | **33.08** | 38.22 |
| SoftMatch weights | 62.56 | 66.92 | 67.71 | 37.59 | 46.43 | 49.09 | 20.61 | 32.29 | 37.13 |
| Weight 1 | 60.68 | 63.97 | 65.02 | 37.36 | 46.37 | 48.95 | 20.30 | 31.73 | 36.49 |
| *Self-training (the model's own predictions)* | | | | | | | | | |
| Squared confidence | 53.80 | 62.59 | 65.12 | 29.44 | 40.39 | 45.44 | 15.78 | 30.05 | 35.98 |
| Fixed $\tau = 0.95$ (FixMatch) | 54.97 | 64.58 | 66.48 | 17.82 | 33.20 | 41.31 | 11.88 | 30.62 | 36.81 |
| FreeMatch threshold | 57.69 | 63.44 | 65.76 | 12.10 | 20.82 | 31.73 | 14.07 | 27.78 | 33.17 |
| SoftMatch weights | 53.05 | 62.99 | 65.31 | 9.86 | 20.53 | 30.14 | 4.79 | 24.16 | 32.93 |
| Weight 1 | 51.17 | 61.52 | 63.95 | 10.71 | 20.25 | 30.71 | 5.63 | 23.93 | 32.12 |
| *Nearest labeled image* | | | | | | | | | |
| Squared confidence | 53.05 | 64.70 | 67.45 | 28.93 | 37.84 | 42.60 | 15.66 | 30.44 | 37.29 |
| Fixed $\tau = 0.95$ | 50.16 | 63.43 | 66.62 | 22.21 | 31.66 | 37.79 | 14.14 | 28.45 | 35.13 |
| FreeMatch threshold | 60.31 | 66.74 | 67.74 | 35.67 | 43.98 | 46.91 | 19.74 | 30.67 | 36.53 |
| SoftMatch weights | 57.90 | 64.61 | 65.67 | 34.51 | 43.07 | 46.19 | 18.97 | 29.91 | 35.53 |
| Weight 1 | 55.96 | 62.66 | 64.06 | 34.51 | 43.09 | 46.19 | 18.89 | 29.82 | 35.42 |

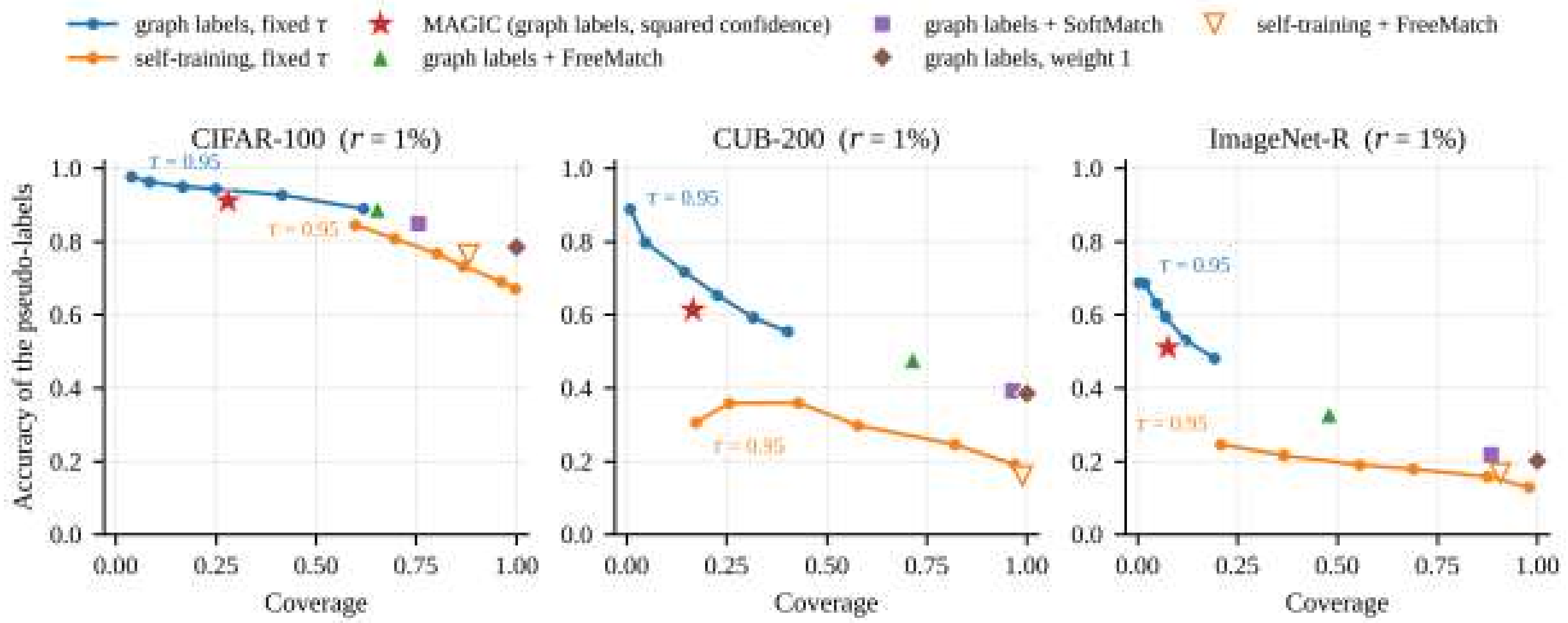

**Figure 3.** *Accuracy of the pseudo-labels used during training against their coverage, $r = 1\%$ (all tasks and the five seeds pooled). Coverage is the mean weight of an unlabeled image; for a threshold, the fraction of images it keeps. Lines: a fixed threshold $\tau$ on the*

*graph's labels (blue) and on the model's own predictions (self-training, orange); the points of a line are $\tau$ = 0.95, 0.9, 0.8, 0.7, 0.5 and 0.3 from left to right. Markers: MAGIC's squared-confidence weights (star) and other gates on the graph's labels and on self-training.*

### 5.5. Ablation study

We now switch MAGIC's parts off one at a time. Every variant is run in all nine settings with the same five seeds as the full model. Table 9 gives the paired differences in AIA; we quote $A_T$ and $F_T$ where they add to the picture.

**Phantom replay is indispensable.** Without it, AIA falls by 37.7–43.1 points on CIFAR-100, 20.5–25.9 on CUB-200 and 11.4–25.3 on ImageNet-R, and forgetting on CIFAR-100 rises to 68.6–83.5 points. The class statistics are useful only because replay turns them into training data for the old classes.

**Most of the gain from unlabeled data flows through the class statistics, not through the loss.** Without any unlabeled data, MAGIC loses 13.2, 14.8 and 5.7 points at $r = 1\%$ on CIFAR-100, CUB-200 and ImageNet-R. When the pseudo-labels are used only in the statistics ($\mathcal{L}_{SSL} = 0$), MAGIC keeps 65%, 74% and 69% of this gain at 1%, and 63–85% across the nine settings. When they are used only in the loss, with statistics from the labeled images, it keeps at most 39%; on CIFAR-100 at 1% it even falls 1.6 points below the model without unlabeled data, and its forgetting rises from 23.2 to 40.0 points. A likely reason is that the loss fits the new classes on many unlabeled images while their stored Gaussians still rest on a few labeled ones, so replay protects them badly in later tasks. The two paths together give the full gain: on top of the statistics, the loss adds 0.7 to 4.6 points. With a fixed threshold of 0.95 instead of the squared-confidence weights, in the loss and in the statistics, at least 97% of the gain is lost in eight of the nine settings, and 72% of it on CIFAR-100 at 1% (Section 5.4).

**The graph's seeds and its second pass matter.** Removing the class prototypes from the first graph (the means of the labeled images and the stored means of the old classes) costs 3.1–3.3 AIA points on CIFAR-100 at every ratio and raises forgetting there by 12.8–17.9 points; on CUB-200 and ImageNet-R its effect stays within 1.2 points. Without the jittered copies of the labeled images (augmented anchors), AIA drops at every ratio on CUB-200 (by 3.6, 3.2 and 1.5 points) and at $r = 1\%$ on CIFAR-100 and ImageNet-R (by 3.1 and 1.5), and changes by less than one point elsewhere. Without the second propagation on the adapted features, whose labels weight the statistics, AIA drops by 0.4–1.7 points in eight of the nine settings.

Ranked by necessity: phantom replay is indispensable; the propagation-weighted statistics carry most of the gain from unlabeled data; the unlabeled loss adds a smaller, consistent gain; the prototype nodes of the graph matter on CIFAR-100 and its augmented anchors at low budgets; the second propagation adds a smaller gain in almost every setting.

**Table 9.** *Ablation of MAGIC in every setting: AIA (%) of each variant minus that of the full model, over the same five seeds (negative: the variant is worse). First row: AIA of the full model. *: the 95% interval of the paired difference excludes zero. Pseudo-labels only in the statistics: $\mathcal{L}_{SSL} = 0$. Pseudo-labels only in the loss: class statistics from the labeled images. Fixed threshold: $\tau = 0.95$ instead of the squared confidence, in the loss and in the statistics. No prototype nodes / no augmented anchors: removed from the first graph. No second propagation: the statistics are weighted with the labels of the first graph.*

| | CIFAR-100 | | | CUB-200 | | | ImageNet-R | | |
|---|---|---|---|---|---|---|---|---|---|
| **Variant** | **1%** | **5%** | **10%** | **1%** | **5%** | **10%** | **1%** | **5%** | **10%** |
| Full MAGIC (AIA) | 63.59 | 68.05 | 69.02 | 37.32 | 45.19 | 47.77 | 19.78 | 32.66 | 38.67 |
| *Where the pseudo-labels act* | | | | | | | | | |
| Labeled images only | −13.19* | −4.11* | −2.47* | −14.84* | −13.47* | −10.03* | −5.71* | −4.21* | −3.24* |
| Only in the statistics | −4.56* | −1.49* | −0.91* | −3.92* | −3.17* | −1.47* | −1.78* | −0.75* | −0.72* |
| Only in the loss | −14.75* | −3.68* | −1.50* | −11.60* | −9.86* | −7.94* | −3.74* | −3.24* | −2.58* |
| Fixed threshold $\tau = 0.95$ | −9.44* | −5.01* | −3.49* | −14.60* | −13.52* | −10.05* | −5.56* | −4.25* | −3.51* |
| *Graph and replay* | | | | | | | | | |
| No prototype nodes | −3.28* | −3.21* | −3.13* | +0.63* | +0.02 | −0.26* | +1.16* | +0.36* | −0.21* |
| No augmented anchors | −3.09* | −0.81* | −0.08 | −3.60* | −3.22* | −1.50* | −1.49* | −0.18 | +0.15 |
| No second propagation | −0.85* | −0.69* | −0.55* | −1.50* | −0.40* | −0.73* | +0.50 | −1.54* | −1.69* |
| Replay off | −37.70* | −42.63* | −43.09* | −20.52* | −24.96* | −25.91* | −11.43* | −22.04* | −25.25* |

## 5.6. Limitations

Our findings should be viewed in light of certain methodological constraints, which we outline below to avoid overstating their general applicability.

(I) All experiments use a pre-trained ImageNet-1K ResNet-18, kept fixed across all methods so that every approach works with a representation of identical capacity. Several recent semi-supervised incremental methods were originally developed on higher-capacity pre-trained vision transformers, which produce considerably more linearly separable features. Sticking with ResNet-18 keeps the cross-method comparison fair, but it also means the relative behavior we observe, in particular the contrast between confidence-threshold selection and manifold-weighted propagation, could shift under a stronger backbone. Consequently, our results should be taken as an extrapolation to the transformer regime, not a direct measurement.

(II) We label the same fraction of every class, so the labeled subset follows the class sizes of the dataset. In practice, label budgets are often long-tailed, and both pseudo-labeling and second-order statistics can degrade unevenly across classes. MAGIC's behavior under imbalanced supervision remains untested. We leave this for future work.

(III) MAGIC stores no raw samples, so it is exemplar-free. It does, however, keep a mean and a diagonal variance per class ($2 \times 512$ numbers with ResNet-18); in that sense it is not statistics-free. Its memory therefore grows with the number of classes and the feature dimension. This footprint is much smaller than that of exemplar replay, but it is not constant and may become a concern at very large class counts.

(IV) The baselines were run with a single class order per dataset, the native one, and with three seeds, each with its own labeled subset. For MAGIC, another class order alone changes the mean

AIA by up to 3.0 points, more than the label split does on CIFAR-100 and ImageNet-R (Appendix C); the baselines may vary as much, so differences of a few points between methods must be read with care. Because the baselines' labeled subsets differ from MAGIC's, our comparisons with them are unpaired, and with three runs their intervals are wide.

(V) MAGIC's unlabeled pathway assumes that same-class images are neighbors in the frozen feature space. On ImageNet-R this holds poorly: at $r = 1\%$, only 21% of the labels that the graph gives the unlabeled images are correct, against 79% on CIFAR-100. The unlabeled stream still adds 3.2 to 5.7 AIA points there (Table 7), but at 5% and 10% labels the methods that adapt the backbone lead by 5 to 7 points. Recovering more would require adapting the representation itself, which the frozen-backbone design forgoes to keep the stored statistics valid; Section 6 outlines a remedy.

(VI) We did not tune MAGIC's gate. A self-adaptive threshold on the graph's labels (FreeMatch) does better than the squared confidence on CUB-200 and on ImageNet-R at 1% (Section 5.4), so MAGIC's gate is not the best possible one. We report MAGIC with the gate fixed before these experiments, rather than choose the gate per dataset from the test results.

## 6. Conclusion

We studied semi-supervised class-incremental learning under scarce supervision, with a frozen backbone and no stored exemplars. In this setting, methods that replay old classes from stored Gaussian statistics face two linked problems: a confidence threshold keeps few or wrong pseudo-labels, and the class statistics on which replay depends are then estimated from a handful of labeled images. MAGIC replaces the threshold by label propagation over a k-nearest-neighbor graph and uses the propagated confidences as weights, both in the unlabeled loss and, through SWGC, in the class statistics, which are kept in the frozen feature space.

With 1% of the labels, MAGIC is the best method on CUB-200, level with the best on ImageNet-R and second on CIFAR-100, 2.2 points behind TACLE; on fine-grained CUB-200 it has the highest mean accuracy at every label ratio (Section 5.2). It gains from unlabeled data in all nine settings, most where labels are scarcest, whereas confidence-thresholded pseudo-labeling attached to supervised methods can help, hurt or do nothing (Section 5.3). Inside MAGIC, the graph's labels are more accurate than the model's own at equal coverage, a fixed threshold of 0.95 removes most of the gain, and most of the gain flows through the class statistics rather than the loss (Sections 5.4 and 5.5). An adaptive threshold on the graph's labels works as well as our weights, and better on CUB-200, so the graph, rather than the exact form of the gate, carries the benefit. At 5–10% labels on CIFAR-100 and ImageNet-R, methods that adapt the backbone overtake MAGIC: its frozen features cannot improve with more labels, and on ImageNet-R they make the graph's labels unreliable.

The main limitation, then, is the frozen backbone. It keeps the stored statistics valid, but at the price of a fixed representation. A natural next step is to let the backbone adapt while keeping the stored statistics consistent with it, through feature distillation or an explicit drift-compensation layer. Further directions are a self-adaptive gate on the graph's labels, pre-trained transformer backbones, and long-tailed label budgets.

**Acknowledgements**

This research is supported by the research grant (No. 2354) of the University of Tabriz.

**Declaration of generative AI use**

During the preparation of this work, the authors used Gemini in order to improve the language and organization of the text and write scripts that compute the statistical tests and plot the figures from the logged experimental results. After using this tool, the authors reviewed and edited the content as needed and take full responsibility for the content of the published article.

**Author contributions: CRediT**

**Yousef Abdi**: Conceptualization, Methodology, Software, Writing - Original Draft.
**Mohammad Asadpour**: Supervision, Conceptualization, Writing - Review & Editing, Validation.
**Yousef Seyfari**: Software, Methodology, Formal analysis, Investigation.

**Code availability**

The code will be made publicly available at https://github.com/yousefabdi/MAGIC-SSCIL upon acceptance.

**Data availability**

All datasets used in this study (CIFAR-100, CUB-200 and ImageNet-R) are publicly available.

## Appendix A: Baseline adaptation protocol

This appendix describes, for each baseline, (i) the method in brief, (ii) where the unlabeled signal enters, (iii) what we changed relative to the original and why, and (iv) the inference rule. Table A1 summarizes and Table A2 lists the full optimization settings.

**LwF [32]**. LwF mitigates forgetting by distillation: a frozen copy of the previous model supplies soft targets ($T = 2$, weight $\lambda_{dist} = 3$) while the backbone is fine-tuned and a linear head is expanded per task. In the semi-supervised setting, a supervised cross-entropy on labeled data is combined with a thresholded FixMatch consistency loss on unlabeled data; the distillation loss requires no labels, so we apply it to labeled and weakly-augmented unlabeled inputs alike. Because LwF stores no prototypes or covariances, no stored statistics can go stale, and standard BatchNorm behavior is retained. Inference is by argmax over the linear head.

**SimpleCIL [62].** SimpleCIL in its original form is training-free: class prototypes are the means of frozen-backbone features, and classification is nearest-prototype under cosine similarity. A parameter-free classifier cannot use unlabeled data, so within the semi-supervised comparison we make the cosine head trainable: labeled-data prototypes initialize the class weights, which FixMatch then refines; the backbone stays frozen and old-class weights are fixed during each new task. The inference rule reported for SimpleCIL is therefore the trained cosine head, not the parameter-free nearest-mean rule. The labeled-only variant recovers the original method exactly.

**APER [62].** APER adapts the backbone once, in the base session, via SSF scale-and-shift parameters, and freezes it afterwards. Its classifier is a set of class prototypes. FixMatch trains the SSF parameters and head during the base session. On every task, prototypes are first estimated from labeled features and then refined with confident pseudo-labeled samples, restricted to the current task's classes. Gradient-based consistency training is confined to the base session because the backbone is frozen thereafter; on later tasks the unlabeled signal enters only through prototype refinement. Inference is nearest-prototype.

**PASS [65]**. PASS retains its full supervised objective (rotation self-supervision, feature distillation, prototype augmentation, all with their published weights $\lambda_{\mathrm{proto}} = \lambda_{\mathrm{fkd}} = 10$) and its published Adam optimizer. Its logit temperature is 1 in all our runs (the official code uses 0.1). Pseudo-labels are computed from weak views on the unrotated logits, restricted to current-task classes. Samples passing the threshold receive the FixMatch consistency loss on their strong views, and PASS's rotation self-supervision is extended to their weak views; feature distillation is also applied to unlabeled weak views. Inference uses the unrotated class logits.

**FeTrIL [41]**. FeTrIL keeps the backbone frozen and represents old classes by translating new features toward stored class means. A linear head is trained on real features of new classes and translated pseudo-features of old ones. FixMatch enters in the head training: pseudo-labels from weak views (restricted to new classes) add confident unlabeled features to the pool, and the consistency loss is applied on strong-view features. Confident pseudo-labeled features also enter the balanced retraining stage alongside the translated features. Departing from the original, which trains the backbone from scratch on a large base session, our matched protocol uses the frozen pre-trained backbone throughout. Inference is by the linear head over frozen features.

**FeCAM [19]**. FeCAM freezes the backbone after the base session and classifies by Mahalanobis distance to class prototypes under per-class shrunk and normalized covariances. FixMatch replaces the supervised objective in base-session backbone training only. Prototype and covariance estimation, and the inference rule, are unchanged, so FixMatch alters only the representation those statistics are computed on.

**RanPAC [38]**. RanPAC projects frozen features through a fixed random matrix ($M = 10{,}000$) with a ReLU and solves a ridge-regression classifier analytically from the accumulators $Q$ (feature–label products) and $G$ (the Gram matrix). The ridge parameter is selected by generalized cross-validation [16] over $\{10^4, \ldots, 10^8\}$ on labeled data. There is no gradient training, so FixMatch here operates on the accumulators: an initial classifier solved from labeled data assigns pseudo-labels to weak views, restricted to current-task classes. Features of confident samples' strong views are then added with confidence weighting, one-hot targets scaled by the confidence in $Q$, and features scaled by its square root in $G$ and the classifier is re-solved. We omit RanPAC's optional first-session parameter-efficient adaptation. The backbone is frozen throughout. Inference is by the analytic linear head.

**SLCA [58]**. SLCA fine-tunes the backbone on every task at a learning rate $100\times$ below the head (the slow learner), accumulates per-class feature Gaussians, and retrains the classifier on features sampled from them (5 epochs, 256 samples per class, logit normalization 0.35). FixMatch operates in the representation stage: a thresholded consistency loss on unlabeled data joins the supervised loss, with pseudo-labels restricted to current-task classes. The classifier-alignment stage and the inference rule are unchanged. BatchNorm runs in eval mode during training (Section 5.1.4).

**LDC [18]**. LDC handles prototype drift by fitting a linear projector that maps old feature-space representations into the current space; classification then becomes a simple nearest-class-mean decision using the drift-compensated prototypes. In our matched protocol we train the backbone with the LwF + FixMatch objective of our LwF baseline: a supervised cross-entropy on labeled images together with FixMatch's consistency loss on the unlabeled ones. Once the backbone is trained, we fit the drift projector on the full image stream (targets are completely ignored) which is a label-free stage with no confidence threshold and running Adam at learning rate $10^{-3}$ for 20 epochs per task. This deliberately departs from the original authors' setup, where the backbone was pre-trained in a self-supervised way (CaSSLe/PFR) for a cold-start scenario. We intentionally keep LDC's projector, which is its central contribution, and plug it into the common pre-trained warm-start backbone so that LDC sits on an equal footing with every other row of the table. At inference time we simply take the nearest class mean over the drift-compensated prototypes.

**USP [11]**. USP is a native SSCIL method built on an iCaRL-style backbone with an exemplar buffer (5,120 samples, herding) and an ETF projection head. Its unlabeled pathway partitions each batch at the confidence threshold: high-confidence samples are pseudo-labeled from the classifier, low-confidence samples are relabeled by nearest class mean in feature space rather than discarded, and a distribution-alignment queue corrects pseudo-label priors. We use its published thresholds and its native 64+64 batch split, 5 warm-up epochs and cosine schedule. Following its original design the buffer is replay-based; USP is the only non-exemplar-free method in the comparison. Inference is by its iCaRL-style head.

**TACLE [27]**. TACLE trains in two stages per task: representation learning with a task-adaptive confidence gate and class-aware weighting of the pseudo-label loss, then classifier alignment on features sampled from stored per-class Gaussians (256 per class). Its threshold follows the inverse-sigmoid schedule $\tau(t) = \alpha/(1 + e^{\alpha t}) + \beta$ with $\alpha = 0.5$, $\beta = 0.65$, decaying from 0.90 toward 0.65 across tasks. Pseudo-labeling begins after a warm-up of one third of the first epoch's iterations. We keep its native 64+64 batch split and stage budgets (14+6 epochs on CIFAR-100). Because stage-2 statistics are reused across tasks, BatchNorm runs in eval mode during training (Section 5.1.4). Inference is by the aligned linear classifier.

**Table A1.** *What each method trains, where unlabeled data enters, and the inference rule. "FT" = fine-tuned.*

| Method | Backbone (base) | Backbone (t>1) | Inference | Unlabeled signal / deviation from original |
|---|---|---|---|---|
| $\text{LwF}_{\text{FM}}$ | FT | FT | Linear head | FixMatch consistency; distillation also on unlabeled weak views. |
| $\text{SimpleCIL}_{\text{FM}}$ | frozen | frozen | Cosine head (trained) | FixMatch trains head from prototype init; original is training-free (recovered by labeled-only variant). |
| $\text{APER}_{\text{FM}}$ | FT (SSF only) | frozen | NCM prototypes | FixMatch trains SSF in base session; prototypes refined with confident pseudo-labels at every task. |
| $\text{PASS}_{\text{FM}}$ | FT | FT | Linear head (unrotated logits) | FixMatch consistency; rotation SSL and distillation extended to confident pseudo-labeled samples. Native Adam; rotation-safe strong augmentation. |
| $\text{FeTrIL}_{\text{FM}}$ | frozen | frozen | Linear head | FixMatch trains the head on frozen features; confident pseudo-features join the translated-feature retraining. Deviation: pre-trained frozen backbone instead of base-session training from scratch. |
| $\text{FeCAM}_{\text{FM}}$ | FT (base only) | frozen | Mahalanobis NCM | FixMatch replaces the supervised base-session objective; statistics and inference unchanged. |
| $\text{RanPAC}_{\text{FM}}$ | frozen | frozen | Analytic ridge head | Confidence-weighted pseudo-label statistics from strong views added to $Q$ and $G$; ridge re-solved. No gradient training; first-session adaptation omitted. |
| $\text{SLCA}_{\text{FM}}$ | FT (slow) | FT (slow) | Aligned linear head | FixMatch in the representation stage; alignment stage unchanged. BatchNorm in eval mode. |
| $\text{LDC}_{\text{FM}}$ | FT | FT | NCM over compensated prototypes | FixMatch consistency on the backbone (weak/strong views), plus LDC's label-free drift projector on the full stream; classification by NCM over drift-compensated prototypes. |
| USP | FT | FT | iCaRL-style head | Native: confidence partitioning with NCM relabeling of low-confidence samples; distribution alignment; buffer 5,120 (herding); $\mu$=1. |
| TACLE | FT | FT | Gaussian-aligned classifier | Native: scheduled threshold 0.90→0.65, class-aware weights, warm-up. BatchNorm in eval mode. |
| MAGIC | frozen | frozen | Adapter + cosine head ($s$=30) | Native: threshold-free label propagation + SWGC; phantom replay from SWGC Gaussians. No confidence gate. |

**Table A2.** *Per-method optimization settings on CIFAR-100 (CUB-200 and ImageNet-R use 40 and 30 epochs per task respectively; other settings unchanged unless noted). Learning rates are head / backbone.*

| Method | Optimizer | LR (head / backbone) | Momentum | Weight decay | LR Schedule | Batch (L+U) | Epoch / task | $\tau$; $T_{\text{sharp}}$ |
|---|---|---|---|---|---|---|---|---|
| $\text{LwF}_{\text{FM}}$ | SGD | $10^{-2}/10^{-4}$ | 0.9 | $5\times10^{-4}$ | Cosine | 16+112 | 20 | 0.95; 1.0 |
| $\text{SimpleCIL}_{\text{FM}}$ | SGD (cosine head) | 0.1/— | 0.9 | $5\times10^{-4}$ | Cosine | 16+112 | 20 | 0.95; 1.0 |
| $\text{APER}_{\text{FM}}$ | AdamW | $5\times10^{-4}$ | — | $5\times10^{-4}$ | Cosine | 16+112 | 20 (base only) | 0.95; 1.0 |
| $\text{PASS}_{\text{FM}}$ | Adam | $5\times10^{-4}/5\times10^{-5}$ | — | $5\times10^{-4}$ | Step at 9, 18 | 16+112 | 20 | 0.95; 1.0 |
| $\text{FeTrIL}_{\text{FM}}$ | SGD (head) | 0.1/— | 0.9 | $5\times10^{-4}$ | Cosine | 16+112 | 20+20 retrain ¶ | 0.95; 1.0 |
| $\text{FeCAM}_{\text{FM}}$ | SGD | $0.1/10^{-3}$ | 0.9 | $5\times10^{-4}$ | Cosine | 16+112 | 20 (base only) | 0.95; 0.05 |
| $\text{RanPAC}_{\text{FM}}$ | closed-form ridge (GCV) | —* | —* | —* | —* | —* | —* | 0.95; 1.0 |
| $\text{SLCA}_{\text{FM}}$ | SGD | $10^{-2}/10^{-4}$ | 0.9 | $5\times10^{-4}$ | MultiStep 45%,60% | 16+112 | 20 + 5 (CA) | 0.95; 1.0 |
| $\text{LDC}_{\text{FM}}$ | SGD | $0.1/10^{-3}$ | 0.9 | $5\times10^{-4}/2\times10^{-4}$ | Step at 7,14,18 (base) / 8, 13 (inc.) | 16+112 | 20 | 0.95; 1.0 |
| projector § | Adam | $10^{-3}$ | — | — | — | 128 (label-free) | 20 | none |
| USP | SGD | $10^{-3}/5\times10^{-4}$ ‖ | 0.9 | $10^{-5}$ | Cosine, 5 ep. wu | 64+64 | 20 | 0.95; cosine + DistAlign |
| TACLE (stage 1) | SGD | $0.02/2\times10^{-4}$ | 0.9 | 0 | Step at 14 | 64+64 | 14 + 6 (CA) | scheduled 0.90→0.65; 1.0 |
| TACLE (stage 2) | SGD (head) | $5\times10^{-3}$ | 0.9 | $5\times10^{-3}$ | Cosine | 256/class | | |
| **MAGIC** | Adam | $10^{-4}$ | — | $10^{-5}$ | constant | 16+112 | 20 | none |

* RanPAC performs no gradient training. Features are extracted in batches of 128 and the classifier is obtained in closed form by ridge regression, with the ridge parameter selected by generalized cross-validation over $\{10^4,\dots,10^8\}$. Learning rate, momentum, weight decay, schedule and epoch budget are therefore inapplicable.

¶ Both of FeTrIL's stages train a linear head on frozen features; the balanced retraining stage over translated pseudo-features is a required part of the method, not an additional budget.

**Table A2.** *Per-method optimization settings on CIFAR-100 (CUB-200 and ImageNet-R use 40 and 30 epochs per task respectively; other settings unchanged unless noted). Learning rates are head / backbone.*

| Method | Optimizer | LR (head / backbone) | Momentum | Weight decay | LR Schedule | Batch (L+U) | Epoch / task | $\tau$; $T_{\text{sharp}}$ |
|---|---|---|---|---|---|---|---|---|

§ LDC consumes the unlabeled stream through two separate loops. Its representation stage pseudo-labels unlabeled images behind a fixed threshold, in 16+112 batches, and updates the backbone. Its drift projector is a distinct loop over the full image stream with the targets discarded: it regresses old-space features onto new-space features ($\mathcal{L} = \| P\Phi_{t-1}(x) - \Phi_t(x) \|^2$) with both backbones frozen, and uses no labels, pseudo-labels or threshold. LDC is thus the only baseline with a label-free unlabeled channel.

‖ USP: the learning rate of the head is $10^{-3}$ in the first task and $5 \times 10^{-4}$ in the later tasks; the learning rate of the backbone is ten times lower.

## Appendix B: Incremental accuracy curves at $r$ = *5%* and *10%*

Per-task accuracy at higher label ratios. Figure B1 extends the $r = 1\%$ curves of Figure 2 to $r = 5\%$ and $10\%$. On fine-grained CUB-200, MAGIC still finishes highest at both ratios (35.9% at 5% and 39.0% at 10%), at least 4.7 points above every other method shown, so its advantage on many-class, few-shot data persists as supervision grows. On CIFAR-100 and ImageNet-R the crossover seen at 1% disappears. MAGIC starts below the backbone-adapting and covariance-based methods (at 10%: 89.1% against 94.4% for SLCA and 94.1% for TACLE on CIFAR-100, and 42.8% against 59.2–61.2% for SLCA, FeCAM and LDC on ImageNet-R) and, although it still loses accuracy more slowly, it no longer overtakes them: it finishes third at 5% and fifth at 10% on both datasets. This is the label-scaling behavior quantified in Section 5.2, and its cause is the starting point, not forgetting: on ImageNet-R at 10%, MAGIC loses 8.4 points from the first task to the last, against 21–27 for SLCA, FeCAM and LDC. On CIFAR-100, TACLE stays within 4 points of SLCA at every task and finishes second at both ratios; on ImageNet-R it ends last of the six methods.

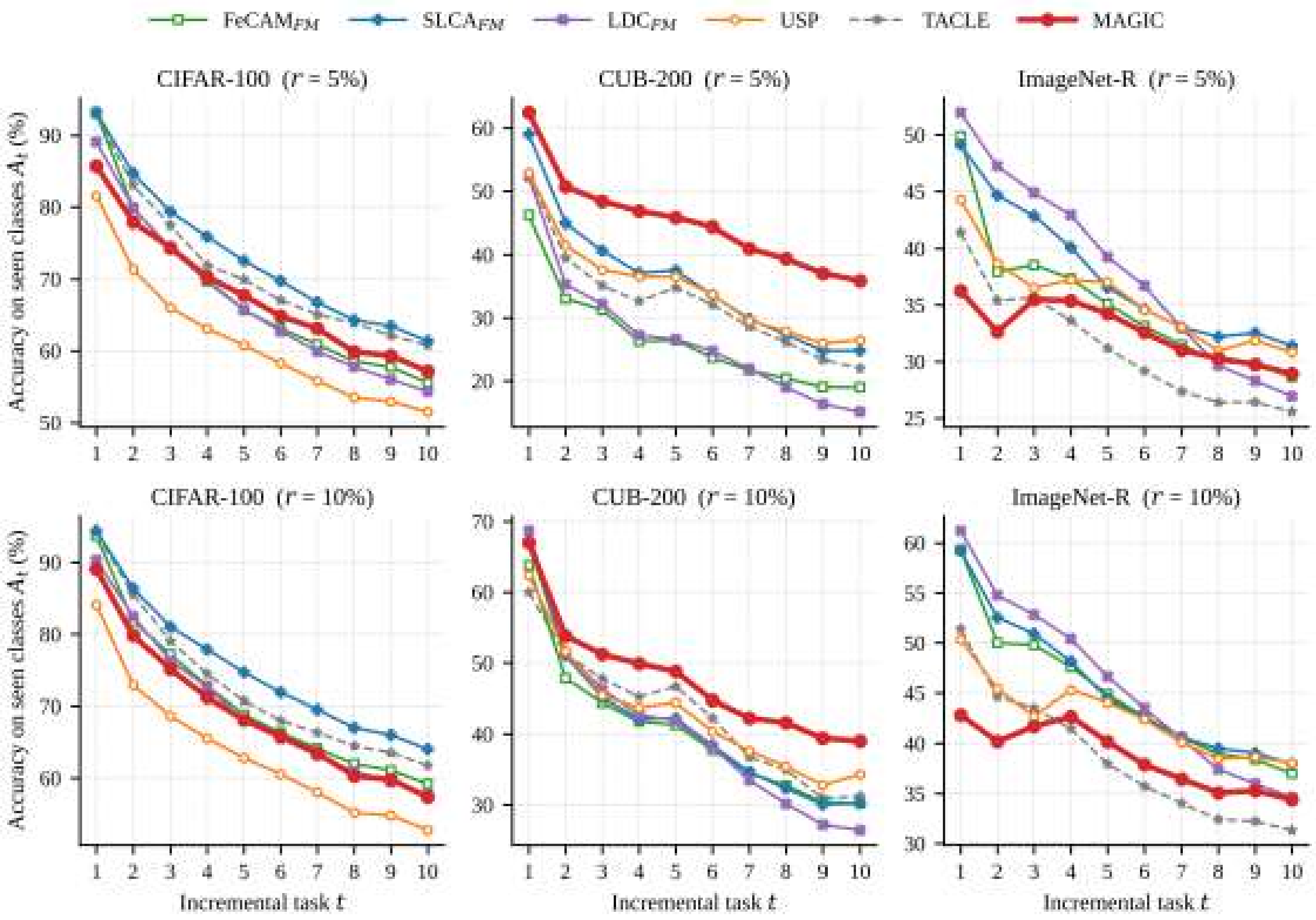

**Figure B1.** *Per-task accuracy $A_t$ (accuracy on all seen classes) for $r = 5\%$ (top row) and $r = 10\%$ (bottom row) across CIFAR-100, CUB-200, and ImageNet-R, averaged over seeds (three for the baselines, five for MAGIC) for six representative methods. Relative to the low-label case in Figure 2 ($r = 1\%$), the trend reverses: as supervision grows, MAGIC no longer overtakes SLCA on CIFAR-100 and ImageNet-R, though on fine-grained CUB-200 it still finishes highest at every label ratio.*

## Appendix C: Sensitivity of MAGIC to the seeds, the label split and the class order

**Seeds.** The baselines were run with three seeds and MAGIC with five, which include those three. Table C1 repeats the comparison of Table 6 with MAGIC restricted to the baselines' seeds (42, 127 and 2026). MAGIC's AIA and last-task accuracy change by at most 0.4 points. The strongest baseline and the sign of the difference are the same in every setting, and the counts of Table 6 change in two of the 18 cases: for AIA on CIFAR-100 at 1%, TACLE is level with MAGIC instead of ahead of it, and for AIA on ImageNet-R at 10%, one of the five baselines that were ahead of MAGIC (RanPAC) becomes level with it.

**Table C1.** *The comparison of Table 6 with MAGIC restricted to the three seeds of the baselines. MAGIC: mean ± std. over these seeds. Δ: MAGIC minus the strongest baseline of Table 6, with the 95% confidence interval of Welch's t-test (three runs each). Ahead / level / behind: as in Table 6.*

| | AIA | | | Last-task accuracy $A_T$ | | |
|---|---|---|---|---|---|---|
| **Labels** | **MAGIC, three seeds** | **Δ [95% CI]** | **Ahead / level / behind** | **MAGIC, three seeds** | **Δ [95% CI]** | **Ahead / level / behind** |
| *CIFAR-100* | | | | | | |
| 1% | 63.85 ± 1.12 | −1.91 [−4.22, +0.40] | 8 / 3 / 0 | 54.51 ± 0.61 | −1.69 [−3.61, +0.23] | 10 / 1 / 0 |
| 5% | 68.10 ± 0.25 | −5.04 [−5.56, −4.51] | 5 / 4 / 2 | 57.08 ± 0.42 | −4.30 [−5.27, −3.34] | 7 / 2 / 2 |
| 10% | 69.06 ± 0.11 | −6.23 [−6.46, −5.99] | 4 / 4 / 3 | 57.41 ± 0.17 | −6.63 [−7.01, −6.25] | 4 / 5 / 2 |
| *CUB-200* | | | | | | |
| 1% | 37.70 ± 1.57 | +9.29 [+3.45, +15.13] | 11 / 0 / 0 | 29.17 ± 0.88 | +9.88 [+6.50, +13.27] | 11 / 0 / 0 |
| 5% | 45.58 ± 1.08 | +5.30 [+1.89, +8.71] | 11 / 0 / 0 | 36.12 ± 0.42 | +7.24 [+5.25, +9.22] | 11 / 0 / 0 |
| 10% | 47.51 ± 1.17 | +1.74 [−1.98, +5.45] | 9 / 2 / 0 | 38.97 ± 1.14 | +4.45 [+2.11, +6.78] | 10 / 1 / 0 |
| *ImageNet-R* | | | | | | |
| 1% | 19.65 ± 0.47 | −0.49 [−1.46, +0.48] | 9 / 2 / 0 | 17.12 ± 0.40 | +1.78 [+0.34, +3.21] | 11 / 0 / 0 |
| 5% | 32.90 ± 1.06 | −5.18 [−7.46, −2.90] | 3 / 6 / 2 | 29.02 ± 0.36 | −2.39 [−3.60, −1.18] | 7 / 3 / 1 |
| 10% | 38.91 ± 0.42 | −6.87 [−9.82, −3.92] | 3 / 4 / 4 | 34.43 ± 0.55 | −3.63 [−5.10, −2.15] | 7 / 1 / 3 |

**Label split and class order.** In all other experiments the seed of a run fixes both the training randomness and the labeled images, and the class order is the native one. To separate these factors, we ran MAGIC with five label splits under each of three class orders, the native order and two random permutations of the classes, that is, 15 runs per setting (Table C2). Over the 15 runs AIA has a standard deviation of 0.7 to 1.9 points, and the 15-run means differ from the five-seed means of Table 3 by at most 1.2 points (CUB-200 at 1%: 38.53 against 37.32). The class order has a systematic effect: with the same label splits, the mean AIA under another order differs from that under the native order by up to 3.0 points (ImageNet-R at 1%: 21.85 against 18.83). The native order is the best of the three on CIFAR-100 and the worst on CUB-200 and ImageNet-R. On CIFAR-100 and ImageNet-R the class order moves AIA more than the label split does (standard deviation 0.85–1.78 against 0.23–0.95 points); on CUB-200 at 1% and 5% the label split matters more.

These numbers calibrate the main comparison. MAGIC's leads over the strongest baseline on CUB-200 at 1% and 5% (8.9 and 4.9 points) and its deficits on CIFAR-100 and ImageNet-R at 5% and 10% (5.1 to 7.1 points) are larger than the shifts caused by the class order; the difference at 1% on ImageNet-R (−0.4 points) is not, and the deficit to TACLE at 1% on CIFAR-100 (2.2 points) is only slightly larger than the largest shift measured there (1.7 points). The baselines were run with the native order only, so the same analysis is not available for them (Section 5.6).

**Table C2.** *MAGIC with five label splits under each of three class orders: AIA (%). All runs: mean ± std. over the 15 runs. Native, random 1, random 2: mean over the five label splits under that class order. Std. over splits: pooled standard deviation over the label splits under the same class order. Std. over orders: pooled standard deviation over the class orders with the same label split.*

| | | Mean under each class order | | | Pooled std. | |
|---|---|---|---|---|---|---|
| **Labels** | **All runs** | **Native** | **Random 1** | **Random 2** | **over splits** | **over orders** |
| *CIFAR-100* | | | | | | |
| 1% | 63.31 ± 0.98 | 63.88 | 62.20 | 63.85 | 0.60 | 1.00 |
| 5% | 67.32 ± 0.83 | 68.01 | 66.24 | 67.72 | 0.23 | 0.96 |
| 10% | 68.23 ± 0.74 | 69.00 | 67.33 | 68.36 | 0.23 | 0.85 |
| *CUB-200* | | | | | | |
| 1% | 38.53 ± 1.90 | 37.43 | 39.55 | 38.62 | 1.81 | 1.35 |
| 5% | 45.40 ± 1.79 | 44.45 | 46.86 | 44.91 | 1.54 | 1.37 |
| 10% | 48.73 ± 1.40 | 47.49 | 50.21 | 48.48 | 0.85 | 1.46 |
| *ImageNet-R* | | | | | | |
| 1% | 20.55 ± 1.59 | 18.83 | 21.85 | 20.99 | 0.95 | 1.78 |
| 5% | 33.53 ± 1.31 | 32.18 | 34.92 | 33.50 | 0.65 | 1.49 |
| 10% | 39.14 ± 1.03 | 38.24 | 40.39 | 38.78 | 0.44 | 1.21 |